\documentclass[runningheads]{llncs}
\usepackage{graphicx}
\usepackage{amsmath,amssymb} 
\usepackage{color}
\usepackage[width=122mm,left=12mm,paperwidth=146mm,height=193mm,top=12mm,paperheight=217mm]{geometry}
\usepackage{multirow}
\usepackage{pbox}

\newcommand{\Ex}[2]{\mathbf{E}_{#1}\left[#2\right]}

\newcommand{\etal}{\textit{et al}}
\begin{document}
\pagestyle{headings}
\mainmatter
\def\ECCV16SubNumber{442}  

\title{Perceptual Losses for Real-Time Style Transfer and Super-Resolution}

\titlerunning{Perceptual Losses for Real-Time Style Transfer and Super-Resolution}

\authorrunning{Johnson \etal}

\author{Justin Johnson, Alexandre Alahi, Li Fei-Fei \\
  {\tt\small\{jcjohns, alahi, feifeili\}@cs.stanford.edu}}
\institute{Department of Computer Science, Stanford University}

\maketitle

\begin{abstract}
We consider image transformation problems, where an input image is transformed into an
output image. Recent methods for such problems typically train feed-forward convolutional
neural networks using a \emph{per-pixel} loss between the output and ground-truth images.
Parallel work has shown that high-quality images can be generated by defining and optimizing
\emph{perceptual} loss functions based on high-level features extracted from pretrained
networks. We combine the benefits of both approaches, and propose the use of perceptual
loss functions for training feed-forward networks for image transformation tasks. We show
results on image
style transfer, where a feed-forward network is trained to solve the optimization
problem proposed by Gatys \etal~in real-time. Compared to the optimization-based method, our
network gives similar qualitative results but is three orders of magnitude faster.
We also experiment with single-image super-resolution, where replacing
a per-pixel loss with a perceptual loss gives visually pleasing results.
\end{abstract}

\section{Introduction}
Many classic problems can be framed as \emph{image transformation} tasks, where a system
receives some input image and transforms it into an output image. Examples from image
processing include denoising, super-resolution, and
colorization, where the input is a degraded image (noisy, low-resolution, or grayscale)
and the output is a high-quality color image. Examples from computer vision
include semantic segmentation and depth estimation, where the input is a color
image and the output image encodes semantic or geometric information about the scene.

One approach for solving image transformation tasks is to train a feed-forward convolutional
neural network in a supervised manner, using a per-pixel loss function to measure the difference
between output and ground-truth images. This approach has been used for example by Dong \etal~for
super-resolution~\cite{dong2015image}, by Cheng \etal~for colorization~\cite{cheng2015deep}, by
Long \etal~for segmentation~\cite{long_shelhamer_fcn}, and by Eigen \etal~for depth and
surface normal prediction~\cite{eigen2014depth,eigen2015predicting}. Such approaches are
efficient at test-time, requiring only a forward pass through the trained network.

However, the per-pixel losses used by these methods do not capture
\emph{perceptual} differences between output and ground-truth images. For example, consider
two identical images offset from each other by one pixel; despite their perceptual similarity
they would be very different as measured by per-pixel losses.

In parallel, recent work has shown that high-quality images can be generated using
\emph{perceptual loss functions} based not on differences between pixels but instead on
differences between high-level image feature representations extracted from pretrained
convolutional neural networks. Images are generated by minimizing a loss function.
This strategy has been applied to feature inversion~\cite{mahendran15understanding}
by Mahendran \etal, to feature visualization by Simonyan \etal~\cite{simonyan2013deep}
and Yosinski \etal~\cite{yosinski2015understanding}, and to texture synthesis and style
transfer by Gatys \etal~\cite{Gatys2015b,gatys2015neural}.  These approaches produce
high-quality images, but are slow since inference requires solving an optimization problem.

\begin{figure}[t]
  \hspace{20mm} Style
  \hspace{15.5mm} Content
  \hspace{5.5mm} Gatys \etal~\cite{gatys2015neural}
  \hspace{6.5mm} Ours
  \vspace{-2.5mm}
  \begin{center}
    \includegraphics[height=0.19\textwidth]{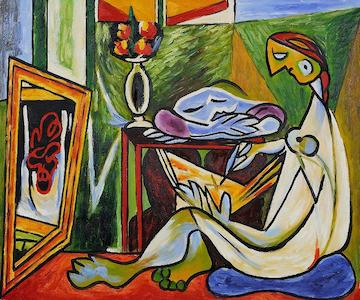}
    \includegraphics[height=0.19\textwidth]{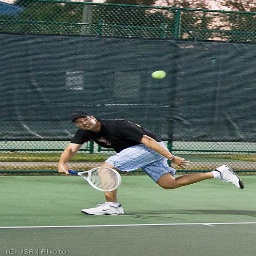}
    \includegraphics[height=0.19\textwidth]{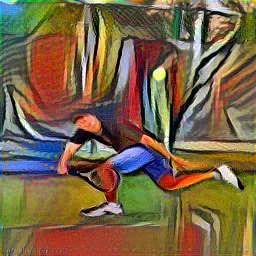}
    \includegraphics[height=0.19\textwidth]{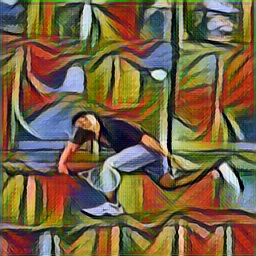} \\
    \vspace{1mm}
    \includegraphics[width=0.19\textwidth]{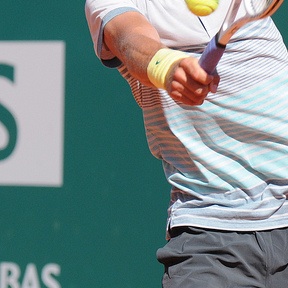}
    \includegraphics[width=0.19\textwidth]{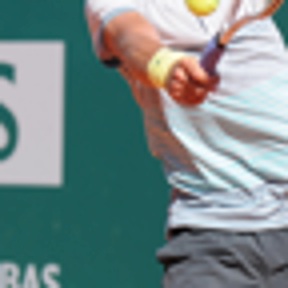}
    \includegraphics[width=0.19\textwidth]{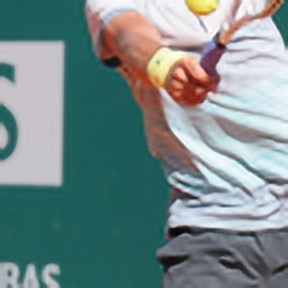}
    \includegraphics[width=0.19\textwidth]{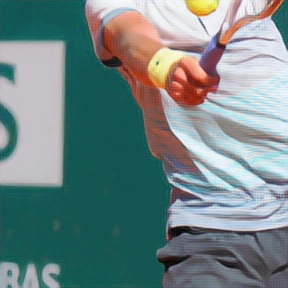}
  \end{center}
  \vspace{-3mm}
  \hspace{14mm} Ground Truth
  \hspace{7mm} Bicubic
  \hspace{8mm} SRCNN~\cite{dong2014learning}
  \hspace{2mm} Perceptual loss
  \vspace{-1mm}
  \caption{Example results for style transfer (top) and $\times 4$ super-resolution
    (bottom). For style transfer, we achieve similar results as Gatys \etal~\cite{gatys2015neural}
    but are three orders of magnitude faster. For super-resolution our method trained with a
    perceptual loss is able to better reconstruct fine details compared to methods trained with
    per-pixel loss.
  }
  \vspace{-5mm}
\end{figure}

In this paper we combine the benefits of these two approaches.
We train feed-forward \emph{transformation networks} for image transformation tasks,
but rather than using \emph{per-pixel} loss functions depending only on low-level pixel
information, we train our networks using \emph{perceptual loss functions} that depend
on high-level features from a pretrained \emph{loss network}. During training, perceptual
losses measure image similarities more robustly than per-pixel losses, and at test-time
the transformation networks run in real-time.

We experiment on two tasks: style transfer and single-image super-resolution. Both are inherently
ill-posed; for style transfer there is no single correct output, and for super-resolution
there are many high-resolution images that could have generated the same low-resolution input.
Success in either task requires semantic reasoning about the input image. For style
transfer the output must be semantically similar to the input despite drastic changes in color
and texture; for super-resolution fine details must be inferred from visually ambiguous low-resolution
inputs. In principle a high-capacity neural network trained for either task could implicitly learn
to reason about the relevant semantics; however in practice we need not learn from scratch: the
use of perceptual loss functions allows the transfer of semantic knowledge from the loss network to
the transformation network.

For style transfer our feed-forward networks are trained to solve the optimization problem from
\cite{gatys2015neural}; our results are similar to \cite{gatys2015neural} both qualitatively and as
measured by objective function value, but are three orders
of magnitude faster to generate. For super-resolution we show that replacing the
per-pixel loss with a perceptual loss gives visually pleasing results for $\times4$ and $\times8$
super-resolution.

\section{Related Work}
\textbf{Feed-forward image transformation.}
In recent years, a wide variety of feed-forward image transformation tasks have been solved
by training deep convolutional neural networks with per-pixel loss functions.

Semantic segmentation methods~\cite{long_shelhamer_fcn,eigen2015predicting,farabet2013learning,pinheiro2013recurrent,noh2015learning,zheng2015conditional}
produce dense scene labels by running a network in a fully-convolutional manner over an input
image, training with a per-pixel classification loss. \cite{zheng2015conditional}
moves beyond per-pixel losses by framing CRF inference as a recurrent layer trained jointly with
the rest of the network. The architecture of our transformation networks are inspired by
\cite{long_shelhamer_fcn} and \cite{noh2015learning}, which use in-network downsampling to reduce
the spatial extent of feature maps followed by in-network upsampling to produce the final
output image.

Recent methods for depth~\cite{eigen2015predicting,eigen2014depth,liu2015deep}
and surface normal estimation~\cite{eigen2015predicting,wang2015designing} are similar in
that they transform a color input image into a geometrically meaningful output image
using a feed-forward convolutional network trained with per-pixel
regression~\cite{eigen2014depth,eigen2015predicting} or classification~\cite{wang2015designing}
losses. Some methods move beyond per-pixel losses by penalizing image
gradients~\cite{eigen2015predicting} or using a CRF loss layer~\cite{liu2015deep} to
enforce local consistency in the output image. In~\cite{cheng2015deep} a feed-forward model
is trained using a per-pixel loss to transform grayscale images to color.

\textbf{Perceptual optimization.}
A number of recent papers have used optimization to generate images where the objective
is perceptual, depending on high-level features extracted from a convolutional network.
Images can be generated to maximize class prediction
scores~\cite{simonyan2013deep,yosinski2015understanding} or individual
features~\cite{yosinski2015understanding} in order to understand the functions encoded in
trained networks. Similar optimization techniques can also be used to generate high-confidence
fooling images~\cite{szegedy2013intriguing,nguyen2015deep}.

Mahendran and Vedaldi~\cite{mahendran15understanding} invert features from convolutional
networks by minimizing a feature reconstruction loss in order to understand the image
information retained by different network layers; similar methods had previously been used
to invert local binary descriptors \cite{d2012beyond} and HOG features
\cite{vondrick2013hoggles}.

The work of Dosovitskiy and Brox~\cite{dosovitskiy2015inverting} is particularly relevant
to ours, as they train a feed-forward neural network to invert convolutional features, quickly
approximating a solution to the optimization problem posed by \cite{mahendran15understanding}.
However, their feed-forward network is trained with a per-pixel reconstruction loss, while
our networks directly optimize the feature reconstruction loss
of~\cite{mahendran15understanding}.

\textbf{Style Transfer.}
Gatys \etal~\cite{gatys2015neural} perform artistic style transfer, combining
the \emph{content} of one image with the \emph{style} of another by jointly minimizing the
feature reconstruction loss of \cite{mahendran15understanding} and a
\emph{style reconstruction loss} also based on features extracted from a pretrained
convolutional network; a similar method had previously been used for texture synthesis~\cite{Gatys2015b}.
Their method produces high-quality results, but is computationally
expensive since each step of the optimization problem requires a forward and backward pass
through the pretrained network. To overcome this computational burden, we train a feed-forward
network to quickly approximate solutions to their optimization problem.

\textbf{Image super-resolution.}
Image super-resolution is a classic problem for which a wide variety of techniques have
been developed. Yang \etal~\cite{yang2014single} provide an exhaustive evaluation of
the prevailing techniques prior to the widespread adoption of convolutional neural networks.
They group super-resolution techniques into prediction-based methods (bilinear, bicubic,
Lanczos, \cite{irani1991improving}), edge-based methods~\cite{freedman2011image,sun2008image},
statistical methods~\cite{shan2008fast,kim2010single,xiong2010robust}, patch-based
methods~\cite{freedman2011image,freeman2002example,chang2004super,glasner2009super,yang2013fast,sun2003image,ni2007image,he2013beta},
and sparse dictionary methods~\cite{yang2008image,yang2010image}.
Recently \cite{dong2015image} achieved excellent performance on single-image
super-resolution using a three-layer convolutional neural network trained with a per-pixel
Euclidean loss. Other recent state-of-the-art methods
include~\cite{timofte2014adjusted,schulter2015fast,huang2015single}.


\begin{figure}[t]
  \centering
  \includegraphics[width=0.98\textwidth]{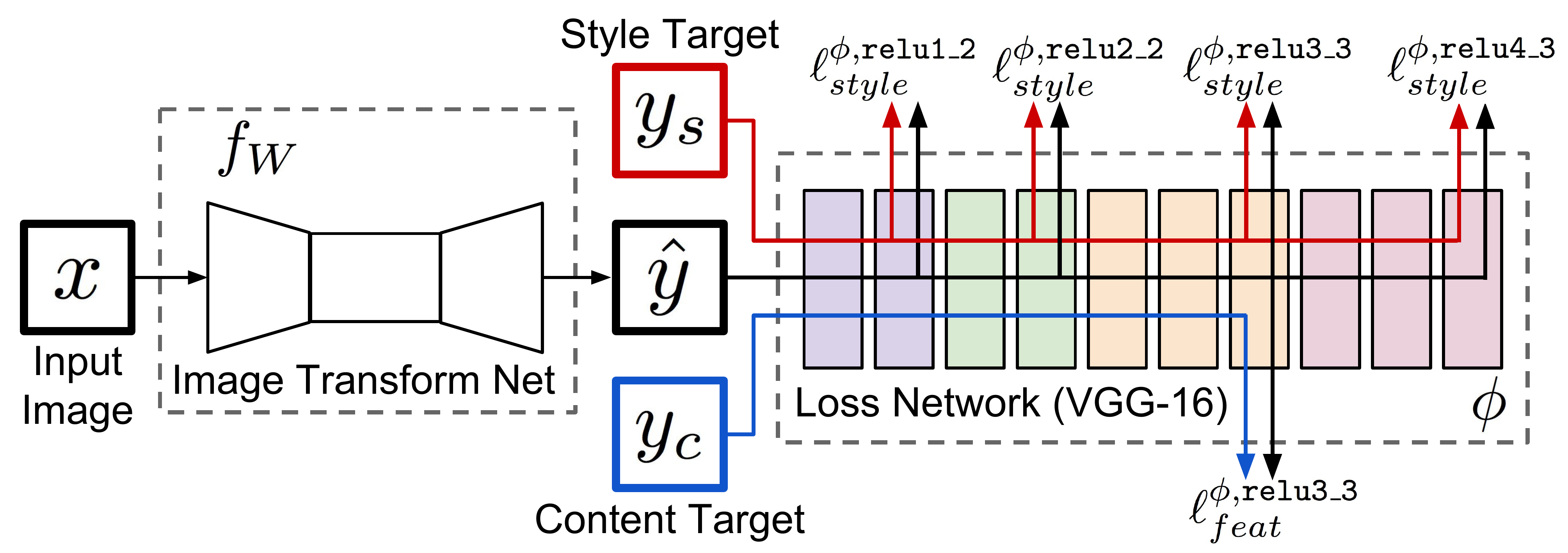}
  \vspace{-3mm}
  \caption{System overview. We train an \emph{image transformation network} to transform
    input images into output images. We use a \emph{loss network} pretrained for image
    classification to define \emph{perceptual loss functions} that measure perceptual
    differences in content and style between images. The loss network remains fixed
    during the training process.
  }
  \vspace{-3mm}
  \label{fig:system}
\end{figure}

\section{Method}
\label{sec:method}
As shown in Figure~\ref{fig:system}, our system consists of two components:
an \emph{image transformation network} $f_W$ and a \emph{loss network} $\phi$ that
is used to define several \emph{loss functions} $\ell_1,\ldots,\ell_k$. The image
transformation network is a deep residual convolutional neural network parameterized by
weights $W$; it transforms input images $x$ into output images $\hat y$ via the
mapping $\hat y = f_W(x)$. Each loss function computes a scalar value $\ell_i(\hat y, y_i)$
measuring the difference between the output image $\hat y$ and a \emph{target image} $y_i$.
The image transformation network is trained using stochastic gradient descent to minimize
a weighted combination of loss functions:

\begin{equation}
  W^* = \arg\min_W \Ex{x, \{y_i\}}{\sum_{i=1} \lambda_i \ell_i(f_W(x), y_i)}
\end{equation}

To address the shortcomings of per-pixel losses and allow our loss functions to better
measure perceptual and semantic differences between images, we draw inspiration from recent
work that generates images via optimization~\cite{mahendran15understanding,simonyan2013deep,yosinski2015understanding,Gatys2015b,gatys2015neural}.
The key insight of these methods is that convolutional neural networks pretrained for image
classification have already learned to encode the perceptual and semantic information we
would like to measure in our loss functions. We therefore make use of a network $\phi$
which as been pretrained for image classification as a fixed \emph{loss network} in order
to define our loss functions. Our deep convolutional transformation network is then trained
using loss functions that are also deep convolutional networks.

The loss network $\phi$ is used to define a \emph{feature reconstruction loss}
$\ell_{feat}^\phi$ and a \emph{style reconstruction loss} $\ell_{style}^\phi$ that measure
differences in content and style between images. For each input image $x$ we have a
\emph{content target} $y_c$ and a \emph{style target} $y_s$. For style transfer, the content
target $y_c$ is the input image $x$ and the output image $\hat y$ should combine the content of
$x=y_c$ with the style of $y_s$; we train one network per style target. For single-image
super-resolution, the input image $x$ is a low-resolution input, the content target $y_c$ is
the ground-truth high-resolution image, and the style reconstruction loss is not used; we
train one network per super-resolution factor.

\subsection{Image Transformation Networks}
\label{sec:arch}
Our image transformation networks roughly follow the architectural guidelines set forth
by Radford \etal~\cite{radford2015unsupervised}. We do not use any pooling layers, instead
using strided and fractionally strided convolutions for in-network downsampling and upsampling.
Our network body consists of five residual blocks~\cite{he2015deep} using the architecture
of~\cite{gross2016training}. All non-residual convolutional layers are followed by spatial batch
normalization~\cite{ioffe2015batch} and ReLU nonlinearities with the exception of the output
layer, which instead uses a scaled $\tanh$ to ensure that the output image has pixels in the
range $[0, 255]$. Other than the first and last layers which use $9\times9$ kernels, all
convolutional layers use $3\times3$ kernels. The exact architectures of all our networks
can be found in the supplementary material.


\vspace{1mm}
\noindent \textbf{Inputs and Outputs.}
For style transfer the input and output are both color images of shape $3\times256\times256$.
For super-resolution with an upsampling factor of $f$, the output is a high-resolution image
patch of shape $3\times288\times288$ and the input is a low-resolution patch of shape
$3\times288/f\times288/f$. Since the image transformation networks are fully-convolutional,
at test-time they can be applied to images of any resolution.

\vspace{1mm}
\noindent \textbf{Downsampling and Upsampling.}
For super-resolution with an upsampling factor of $f$, we use several residual
blocks followed by $\log_2f$ convolutional layers with stride 1/2. This is different from
\cite{dong2015image} who use bicubic interpolation to upsample the low-resolution input
before passing it to the network. Rather than relying on a fixed upsampling function,
fractionally-strided convolution allows the upsampling function to be learned jointly with
the rest of the network.

For style transfer our networks use two stride-2 convolutions to downsample the input
followed by several residual blocks and then two convolutional layers with stride 1/2
to upsample. Although the input and output have the same size, there are several benefits to
networks that downsample and then upsample.

The first is computational. With a naive implementation, a $3\times3$ convolution with $C$ filters on
an input of size $C\times H\times W$ requires $9HWC^2$ multiply-adds, which is the same cost
as a $3\times3$ convolution with $DC$ filters on an input of shape $DC\times H/D\times W/D$.
After downsampling, we can therefore use a larger network for the same computational cost.

The second benefit has to do with effective receptive field sizes. High-quality style
transfer requires changing large parts of the image in a coherent way; therefore it is
advantageous for each pixel in the output to have a large effective receptive field in
the input. Without downsampling, each additional $3\times3$ convolutional layer
increases the effective receptive field size by 2. After downsampling by a factor of
$D$, each $3\times3$ convolution instead increases effective receptive field size by $2D$,
giving larger effective receptive fields with the same number of layers.

\vspace{1mm}
\noindent \textbf{Residual Connections.}
He \etal~\cite{he2015deep} use \emph{residual connections} to train very deep
networks for image classification. They argue that residual connections make it easy for the
network to learn the identify function; this is an appealing property for image transformation
networks, since in most cases the output image should share structure with the input image.
The body of our network thus consists of several residual blocks, each of which contains
two $3\times3$ convolutional layers. We use the residual block design
of~\cite{gross2016training}, shown in the supplementary material.


\begin{figure}[t]
  \hspace{10.5mm} $y$
  \hspace{10mm} \verb|relu2_2|
  \hspace{5.5mm} \verb|relu3_3|
  \hspace{5.5mm} \verb|relu4_3|
  \hspace{5.5mm} \verb|relu5_1|
  \hspace{5.5mm} \verb|relu5_3|
  \vspace{-2mm}
  \begin{center}
    \includegraphics[width=0.15\textwidth]{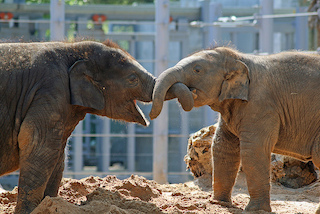}
    \includegraphics[width=0.15\textwidth]{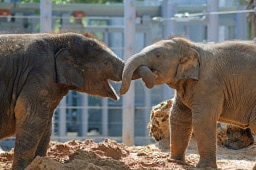}
    \includegraphics[width=0.15\textwidth]{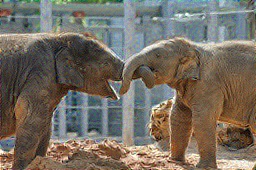}
    \includegraphics[width=0.15\textwidth]{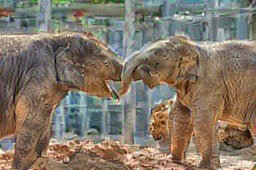}
    \includegraphics[width=0.15\textwidth]{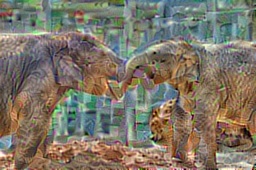}
    \includegraphics[width=0.15\textwidth]{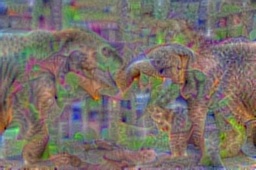} \\
    \includegraphics[width=0.15\textwidth]{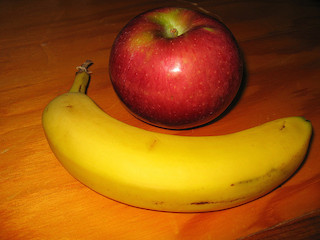}
    \includegraphics[width=0.15\textwidth]{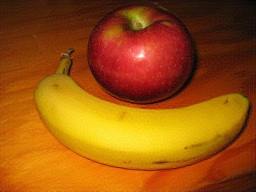}
    \includegraphics[width=0.15\textwidth]{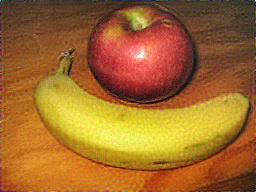}
    \includegraphics[width=0.15\textwidth]{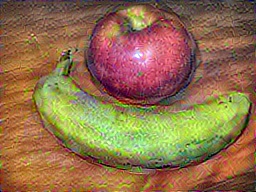}
    \includegraphics[width=0.15\textwidth]{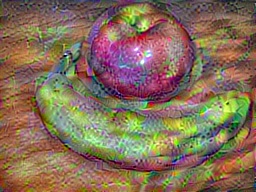}
    \includegraphics[width=0.15\textwidth]{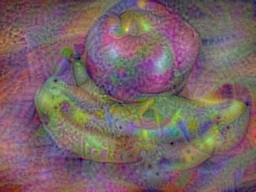}
  \end{center}
  \vspace{-5mm}
  \caption{Similar to \cite{mahendran15understanding}, we use optimization to find an image
    $\hat y$ that minimizes the feature reconstruction loss
    $\ell_{feat}^{\phi, j}(\hat y, y)$ for several layers $j$ from the pretrained VGG-16
    loss network $\phi$. As we reconstruct from higher layers, image content and overall
    spatial structure are preserved, but color, texture, and exact shape are not.
  }
  \label{fig:feature-loss}
\end{figure}

\subsection{Perceptual Loss Functions}
We define two \emph{perceptual loss functions} that measure high-level perceptual and semantic
differences between images. They make use of a \emph{loss network} $\phi$ pretrained
for image classification, meaning that these perceptual loss functions
are themselves deep convolutional neural networks. In all our experiments $\phi$ is the 
16-layer VGG network~\cite{simonyan2014very} pretrained on the ImageNet dataset~\cite{ILSVRC15}.

\vspace{1mm}
\noindent\textbf{Feature Reconstruction Loss.}
Rather than encouraging the pixels of the output image $\hat y=f_W(x)$ to exactly match
the pixels of the target image $y$, we instead encourage them to have similar feature
representations as computed by the loss network $\phi$. Let $\phi_j(x)$ be the activations
of the $j$th layer of the network $\phi$ when processing the image $x$; if $j$ is a
convolutional layer then $\phi_j(x)$ will be a feature map of shape
$C_j \times H_j \times W_j$. The \emph{feature reconstruction loss} is
the (squared, normalized) Euclidean distance between feature representations:
\begin{equation}
  \ell_{feat}^{\phi,j}(\hat y, y) = 
  \frac1{C_jH_jW_j}\|\phi_j(\hat y) - \phi_j(y)\|_2^2
\end{equation}
As demonstrated in \cite{mahendran15understanding} and reproduced in
Figure~\ref{fig:feature-loss}, finding an image $\hat y$ that minimizes the feature
reconstruction loss for early layers tends to produce images that are visually
indistinguishable from $y$. As we reconstruct from higher layers, image content
and overall spatial structure are preserved but color, texture, and exact shape are not.
Using a feature reconstruction loss for training our image transformation networks encourages
the output image $\hat y$ to be perceptually similar to the target image $y$, but does not
force them to match exactly.


\begin{figure}[t]
  \hspace{11mm} $y$
  \hspace{15mm} \verb|relu1_2|
  \hspace{10mm} \verb|relu2_2|
  \hspace{10mm} \verb|relu3_3|
  \hspace{10mm} \verb|relu4_3|
  \vspace{-2.5mm}
  \begin{center}
    \raisebox{1.5mm}{\includegraphics[width=0.19\textwidth]{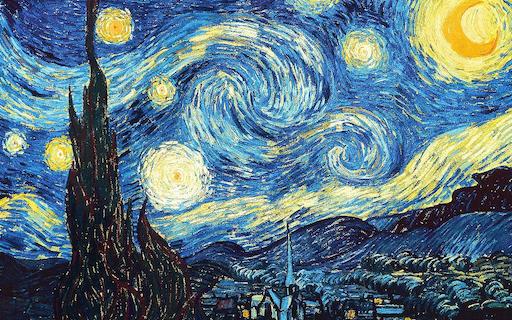}}
    \includegraphics[width=0.19\textwidth]{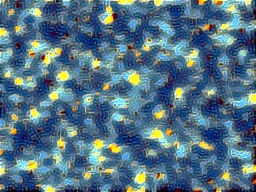}
    \includegraphics[width=0.19\textwidth]{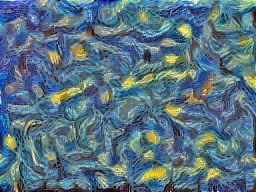}
    \includegraphics[width=0.19\textwidth]{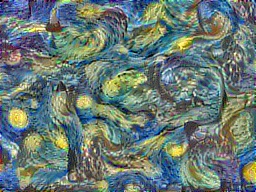}
    \includegraphics[width=0.19\textwidth]{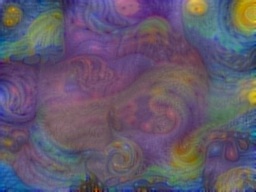} \\
    \includegraphics[width=0.19\textwidth]{la_muse_small.jpg}
    \includegraphics[width=0.19\textwidth]{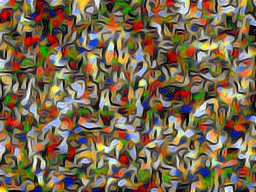}
    \includegraphics[width=0.19\textwidth]{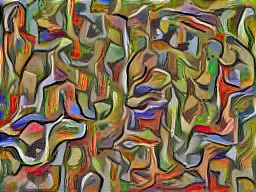}
    \includegraphics[width=0.19\textwidth]{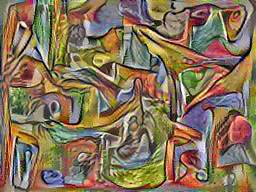}
    \includegraphics[width=0.19\textwidth]{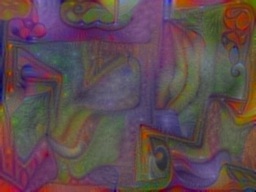}
  \end{center}
  \vspace{-5mm}
  \caption{Similar to~\cite{gatys2015neural}, we use optimization to find an image $\hat y$
    that minimizes the style reconstruction loss $\ell_{style}^{\phi, j}(\hat y, y)$
    for several layers $j$ from the pretrained VGG-16 loss network $\phi$. The images
    $\hat y$ preserve stylistic features but not spatial structure.
  }
  \label{fig:style-loss}
\end{figure}

\vspace{1mm}
\noindent\textbf{Style Reconstruction Loss.}
The feature reconstruction loss penalizes the output image $\hat y$ when it deviates in
content from the target $y$. We also wish to penalize differences in style:
colors, textures, common patterns, etc. To achieve this effect, Gatys
\etal~\cite{Gatys2015b,gatys2015neural} propose the following \emph{style reconstruction loss}.

As above, let $\phi_j(x)$ be the activations at the $j$th layer of the network $\phi$
for the input $x$, which is a feature map of shape
$C_j\times H_j\times W_j$. Define the \emph{Gram matrix} $G^\phi_j(x)$ to be the
$C_j\times C_j$ matrix whose elements are given by

\begin{equation}
  G^\phi_j(x)_{c, c'} = \frac1{C_jH_jW_j}\sum_{h=1}^{H_j}\sum_{w=1}^{W_j}\phi_j(x)_{h,w,c}\phi_j(x)_{h,w,c'}.
\end{equation}

If we interpret $\phi_j(x)$ as giving $C_j$-dimensional features for each point on a
$H_j\times W_j$ grid, then $G^\phi_j(x)$ is proportional to the uncentered covariance of the
$C_j$-dimensional features, treating each grid location as an independent sample.
It thus captures information about which features tend to activate together.
The Gram matrix can be computed efficiently by reshaping $\phi_j(x)$ into a matrix $\psi$ of
shape $C_j\times H_jW_j$; then $G^\phi_j(x) = \psi\psi^T/C_jH_jW_j$.

The \emph{style reconstruction loss} is then the squared Frobenius norm of the difference between
the Gram matrices of the output and target images:
\begin{equation}
  \ell_{style}^{\phi, j}(\hat y, y) = \|G^\phi_j(\hat y) - G^\phi_j(y)\|_F^2.
\end{equation}
The style reconstruction loss is well-defined even when $\hat y$ and $y$ have different
sizes, since their Gram matrices will both have the same shape.

As demonstrated in \cite{gatys2015neural} and reproduced in Figure~\ref{fig:style-loss},
generating an image $\hat y$ that minimizes the style reconstruction loss preserves stylistic
features from the target image, but does not preserve its spatial structure. Reconstructing
from higher layers transfers larger-scale structure from the target image.

To perform style reconstruction from a set of layers $J$ rather than a single layer $j$,
we define $\ell_{style}^{\phi, J}(\hat y, y)$ to be the sum of losses for each layer $j\in J$.

\subsection{Simple Loss Functions}
In addition to the perceptual losses defined above, we also define two simple loss functions
that depend only on low-level pixel information.

\vspace{1mm}
\noindent \textbf{Pixel Loss.}
The \emph{pixel loss} is the (normalized) Euclidean distance between the output image
$\hat y$ and the target $y$. If both have shape $C\times H\times W$, then the pixel loss is
defined as $\ell_{pixel}(\hat y, y) = \|\hat y - y\|^2_2 / CHW$. This can only be used
when when we have a ground-truth target $y$ that the network is expected to match.

\vspace{1mm}
\noindent\textbf{Total Variation Regularization.}
To encourage spatial smoothness in the output image $\hat y$, we
follow prior work on feature inversion \cite{mahendran15understanding,d2012beyond}
and super-resolution~\cite{aly2005image,zhang2010non} and make use of
\emph{total variation regularizer} $\ell_{TV}(\hat y)$.

\section{Experiments}
We perform experiments on two image transformation tasks: style transfer and single-image
super-resolution. Prior work on style transfer has used optimization to generate images;
our feed-forward networks give similar qualitative results but are up to three orders of
magnitude faster. Prior work on single-image super-resolution with convolutional neural
networks has used a per-pixel loss; we show encouraging qualitative results by using
a perceptual loss instead.

\subsection{Style Transfer}
\label{sec:style}
The goal of style transfer is to generate an image $\hat y$ that combines the content of
a \emph{target content image} $y_c$ with the the \emph{style} of a \emph{target style image}
$y_s$. We train one image transformation network per style target for several hand-picked
style targets and compare our results with the baseline approach of
Gatys \etal~\cite{gatys2015neural}.

\begin{figure}[t]
  \centering
  \includegraphics[width=0.9\textwidth]{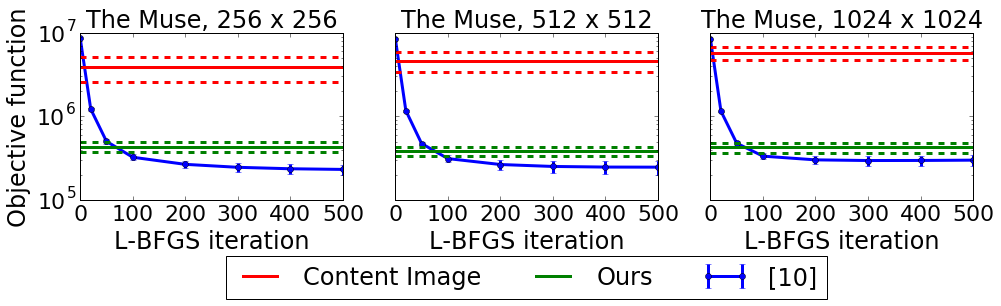}
  \vspace{-4mm}
  \caption{Our style transfer networks and \cite{gatys2015neural} minimize the same objective.
    We compare their objective values on 50 images; dashed lines and error bars show standard
    deviations. Our networks are trained on $256\times256$ images but generalize to larger images.
  }
  \label{fig:style-loss}
  \vspace{-6mm}
\end{figure}

\vspace{1mm}
\noindent \textbf{Baseline.}
As a baseline, we reimplement the method of Gatys \etal~\cite{gatys2015neural}. Given style
and content targets $y_s$ and $y_c$ and layers $j$ and $J$ at which to perform feature and
style reconstruction, an image $\hat y$ is generated by solving the problem
\begin{equation}
  \hat y = \arg\min_{y} \lambda_c \ell_{feat}^{\phi,j}(y, y_c)
    + \lambda_s\ell_{style}^{\phi,J}(y, y_s) + \lambda_{TV} \ell_{TV}(y)
  \label{eq:style}
\end{equation}
where $\lambda_c,\lambda_s$, and $\lambda_{TV}$ are scalars, $y$ is initialized with white
noise, and optimization is performed using L-BFGS. We find that unconstrained optimization of
Equation~\ref{eq:style} typically results in images whose pixels fall outside the range
$[0, 255]$. For a more fair comparison with our method whose output is constrained to this
range, for the baseline we minimize Equation~\ref{eq:style} using projected L-BFGS by
clipping the image $y$ to the range $[0, 255]$ at each iteration. In most cases optimization
converges to satisfactory results within 500 iterations. This method is slow because each
L-BFGS iteration requires a forward and backward pass through the VGG-16 loss network $\phi$. 

\vspace{1mm}
\noindent \textbf{Training Details.}
Our style transfer networks are trained on the Microsoft COCO dataset~\cite{lin2014microsoft}.
We resize each of the 80k training images to $256\times256$ and train our networks with
a batch size of 4 for 40,000 iterations, giving roughly two epochs over the training data.
We use Adam~\cite{kingma2014adam} with a learning rate of $1\times10^{-3}$. The output images
are regularized with total variation regularization with a strength of between $1\times10^{-6}$
and $1\times10^{-4}$, chosen via cross-validation per style target. We do not use weight decay
or dropout, as the model does not overfit within two epochs. For all style transfer experiments
we compute feature reconstruction loss at layer \verb.relu2_2. and style reconstruction
loss at layers \verb.relu1_2., \verb.relu2_2., \verb.relu3_3., and \verb.relu4_3. of the
VGG-16 loss network $\phi$.
Our implementation uses Torch~\cite{collobert2011torch7} and cuDNN~\cite{chetlur2014cudnn};
training takes roughly 4 hours on a single GTX Titan X GPU.

\begin{figure}
  \raisebox{7mm}{
    \begin{minipage}{27mm}
      \centering
      \textbf{Style} \\
      \textit{The Starry Night},
      Vincent van Gogh, 1889
    \end{minipage}
  }
  \includegraphics[height=0.13\textwidth]{starry_night_small.jpg}
  \hspace{7mm}
  \raisebox{7mm}{
    \begin{minipage}{25mm}
      \centering
      \textbf{Style} \\
      \textit{The Muse},\\
      Pablo Picasso, 1935
    \end{minipage}
  }
  \includegraphics[height=0.13\textwidth]{la_muse_small.jpg} \\
  \includegraphics[width=0.16\textwidth]{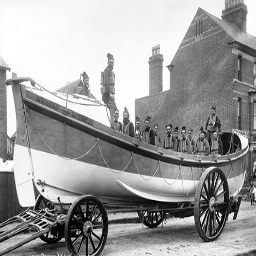}
  \includegraphics[width=0.16\textwidth]{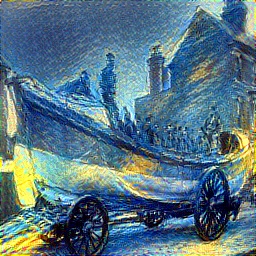}
  \includegraphics[width=0.16\textwidth]{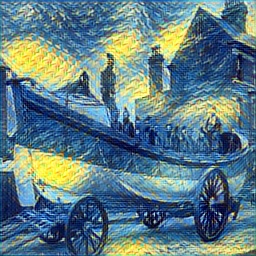}
  \includegraphics[width=0.16\textwidth]{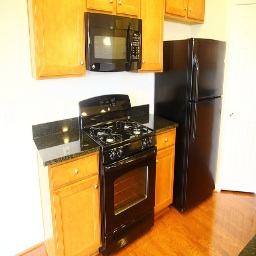}
  \includegraphics[width=0.16\textwidth]{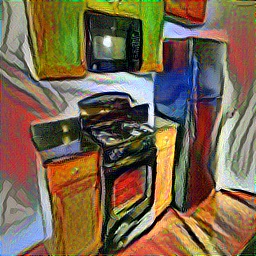}
  \includegraphics[width=0.16\textwidth]{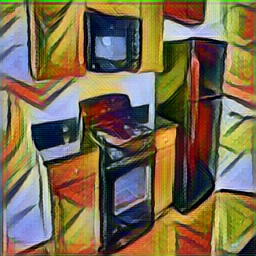} \\
  \includegraphics[width=0.16\textwidth]{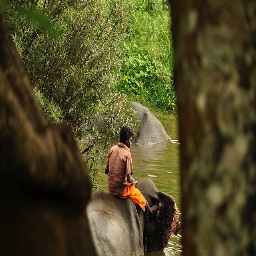}
  \includegraphics[width=0.16\textwidth]{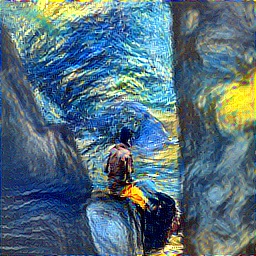}
  \includegraphics[width=0.16\textwidth]{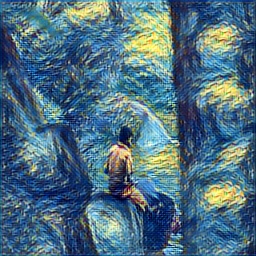}
  \includegraphics[width=0.16\textwidth]{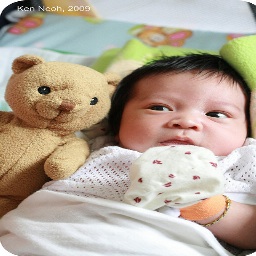}
  \includegraphics[width=0.16\textwidth]{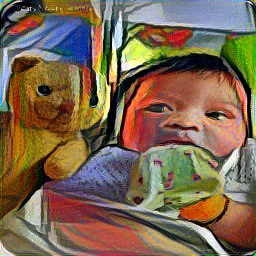}
  \includegraphics[width=0.16\textwidth]{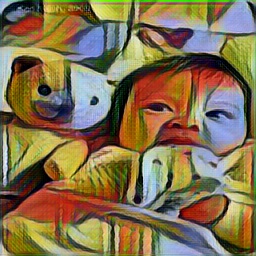} \\
  \raisebox{7mm}{
    \begin{minipage}{27mm}
      \centering
      \textbf{Style} \\
      \textit{Composition VII},
      Wassily Kandinsky, 1913
    \end{minipage}
  }
  \includegraphics[height=0.13\textwidth]{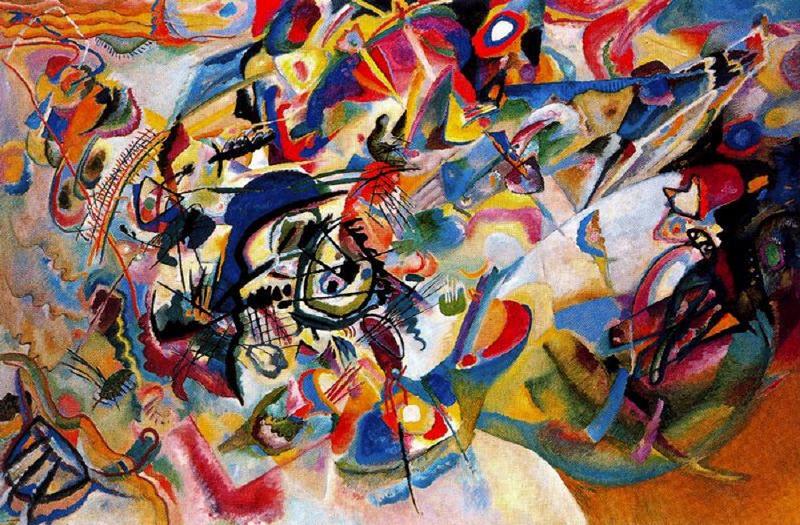}
  \hspace{1mm}
  \raisebox{7mm}{
    \begin{minipage}{30mm}
      \centering
      \textbf{Style} \\
      \textit{The Great Wave off Kanagawa},
      Hokusai, 1829-1832
    \end{minipage}
  }
  \includegraphics[height=0.13\textwidth]{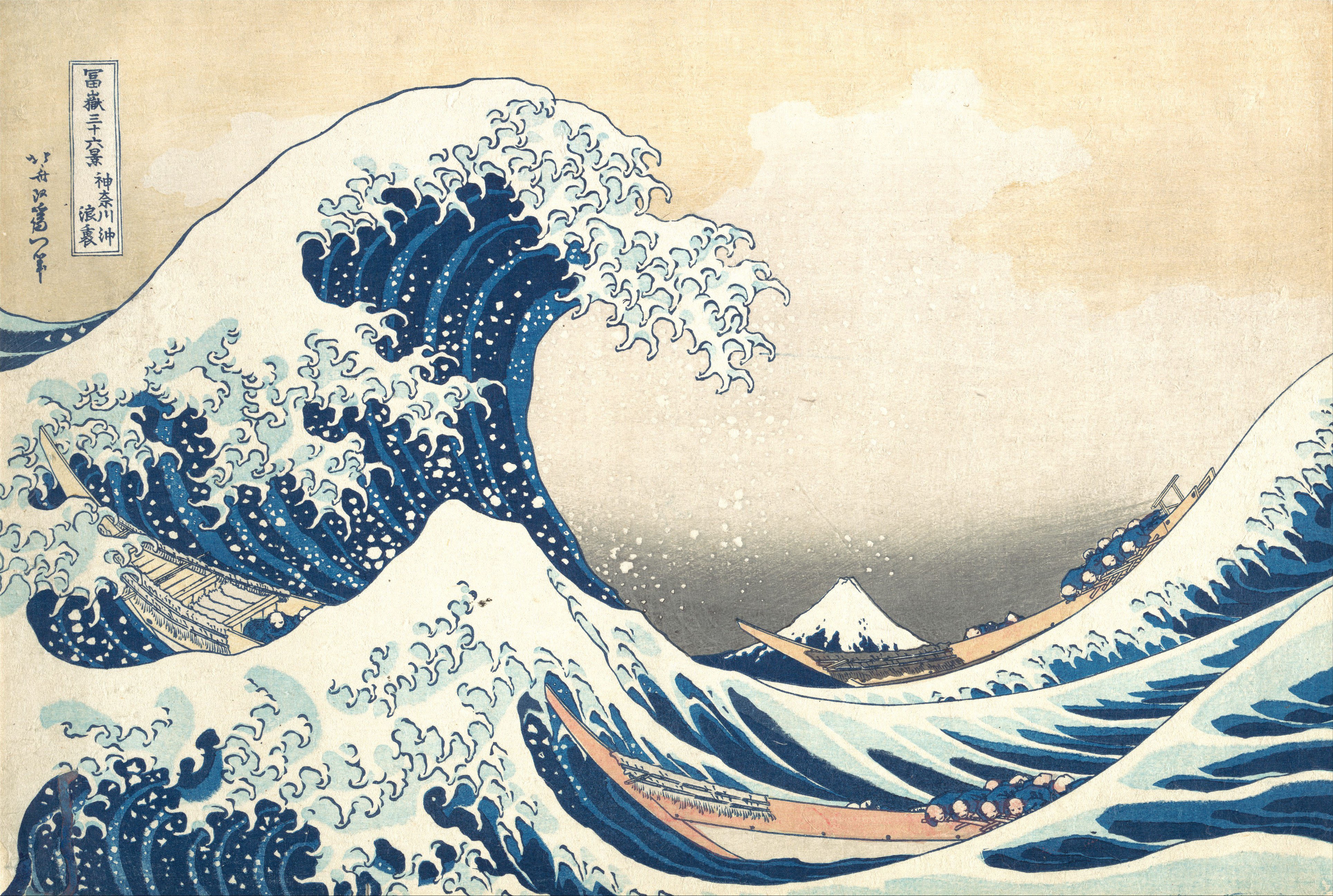} \\
  \includegraphics[width=0.16\textwidth]{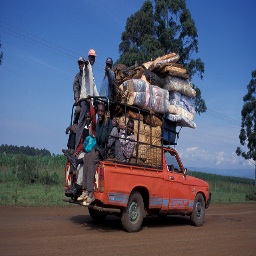}
  \includegraphics[width=0.16\textwidth]{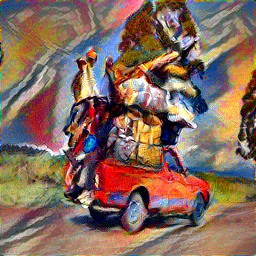}
  \includegraphics[width=0.16\textwidth]{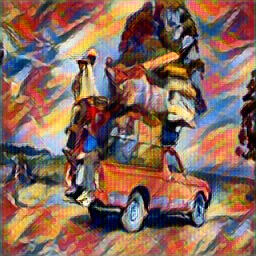}
  \includegraphics[width=0.16\textwidth]{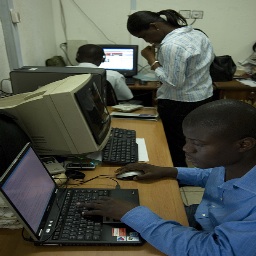}
  \includegraphics[width=0.16\textwidth]{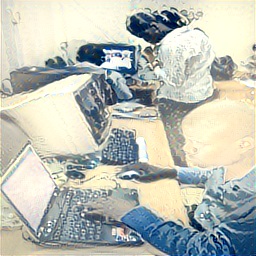}
  \includegraphics[width=0.16\textwidth]{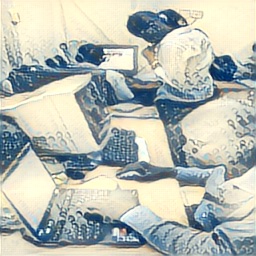} \\
  \includegraphics[width=0.16\textwidth]{34_orig.jpg}
  \includegraphics[width=0.16\textwidth]{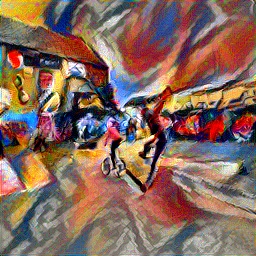}
  \includegraphics[width=0.16\textwidth]{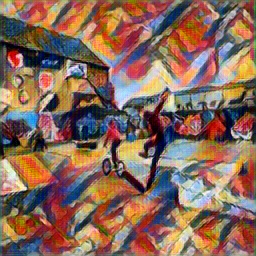}
  \includegraphics[width=0.16\textwidth]{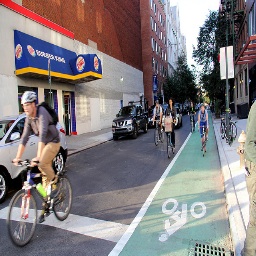}
  \includegraphics[width=0.16\textwidth]{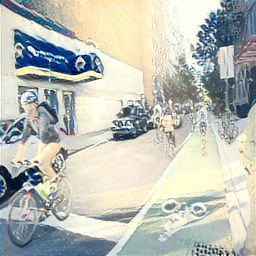}
  \includegraphics[width=0.16\textwidth]{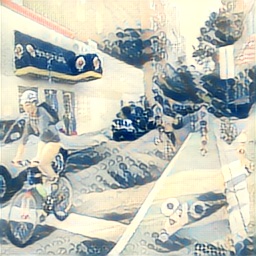} \\
  \hspace*{12mm}
  \raisebox{7mm}{
    \begin{minipage}{16mm}
      \centering
      \textbf{Style} \\
      \textit{Sketch}
    \end{minipage}
  }
  \includegraphics[height=0.13\textwidth]{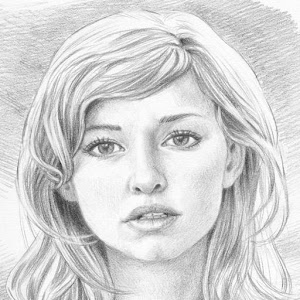}
  \hspace{8mm}
  \raisebox{7mm}{
    \begin{minipage}{25mm}
      \centering
      \textbf{Style} \\
      \textit{The Simpsons}
    \end{minipage}
  }
  \includegraphics[height=0.13\textwidth]{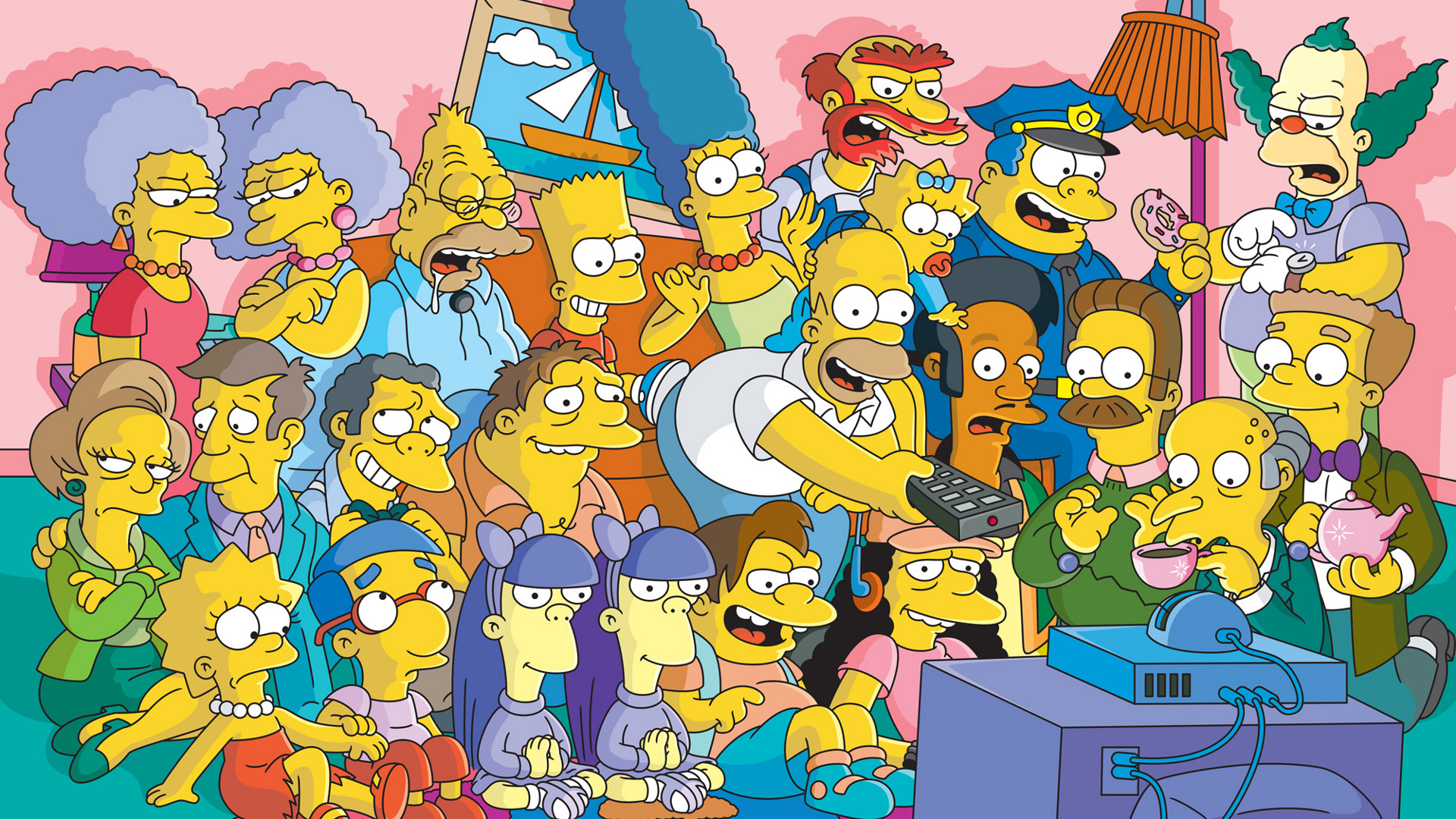} \\
  \includegraphics[width=0.16\textwidth]{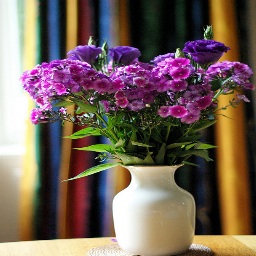}
  \includegraphics[width=0.16\textwidth]{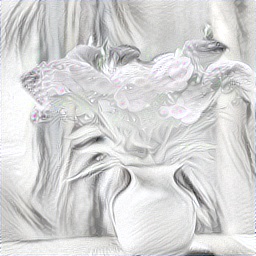}
  \includegraphics[width=0.16\textwidth]{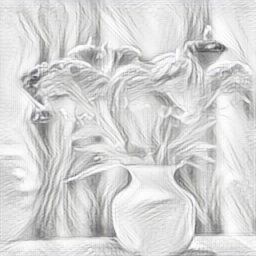}
  \includegraphics[width=0.16\textwidth]{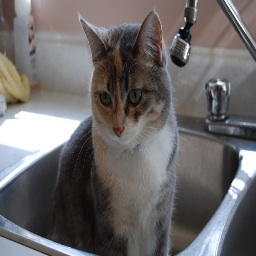}
  \includegraphics[width=0.16\textwidth]{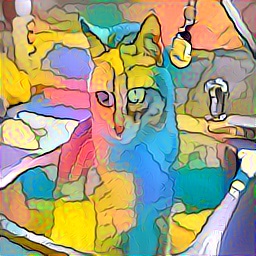}
  \includegraphics[width=0.16\textwidth]{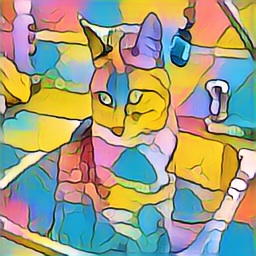} \\
  \includegraphics[width=0.16\textwidth]{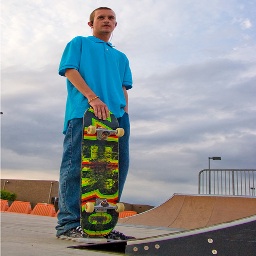}
  \includegraphics[width=0.16\textwidth]{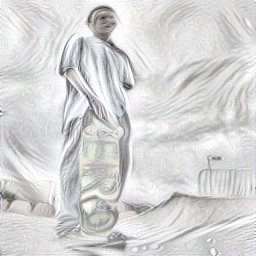}
  \includegraphics[width=0.16\textwidth]{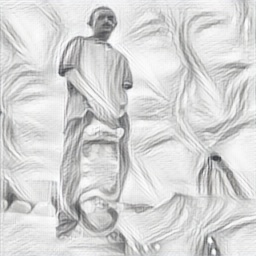} 
  \includegraphics[width=0.16\textwidth]{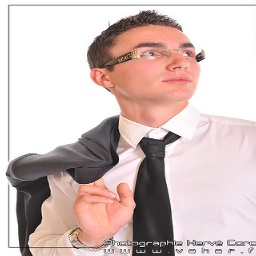}
  \includegraphics[width=0.16\textwidth]{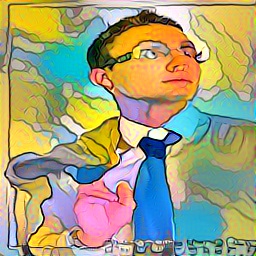}
  \includegraphics[width=0.16\textwidth]{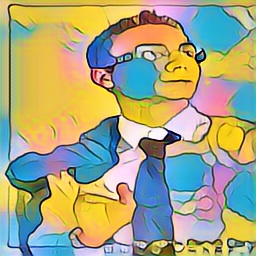} \\
  \hspace*{2.5mm} \textbf{Content}
  \hspace{9mm} \cite{gatys2015neural}
  \hspace{12mm} Ours
  \hspace{9.5mm} \textbf{Content}
  \hspace{9mm} \cite{gatys2015neural}
  \hspace{12mm} Ours \\
  \vspace{-5mm}
  \caption{Example results of style transfer using our image transformation networks.
    Our results are qualitatively similar to Gatys \etal~\cite{gatys2015neural}
    but are much faster to generate (see Table~\ref{tab:style-speed}).
    All generated images are $256\times256$ pixels.
  }
  \label{fig:style-results}
\end{figure}

\vspace{1mm}
\noindent \textbf{Qualitative Results.}
In Figure~\ref{fig:style-results} we show qualitative examples comparing our results with those
of the baseline method for a variety of style and content images. In all cases the hyperparameters
$\lambda_c$, $\lambda_s$, and $\lambda_{TV}$ are exactly the same between the two methods; all
content images are taken from the MS-COCO 2014 validation set. Overall our results
are qualitatively similar to the baseline.

\begin{figure}[t]
  \centering
  \includegraphics[width=0.21\textwidth]{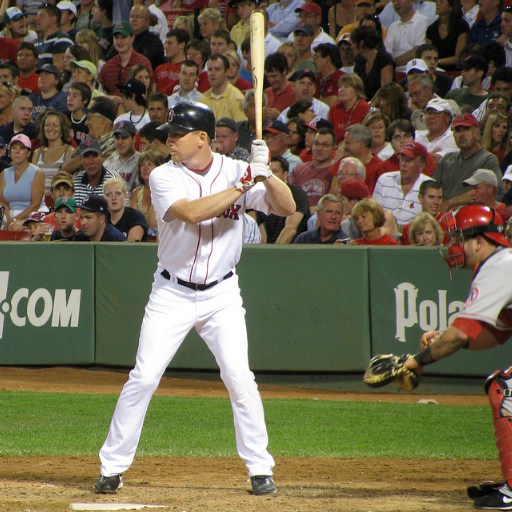} 
  \includegraphics[width=0.21\textwidth]{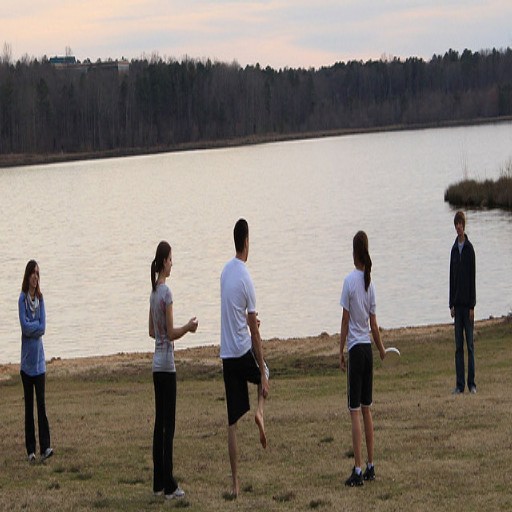}
  \includegraphics[width=0.21\textwidth]{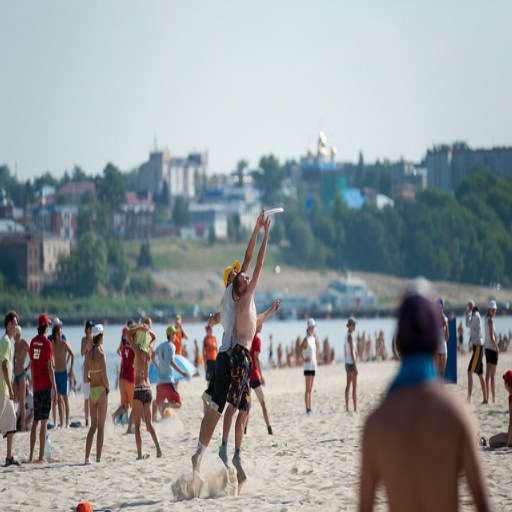}
  \includegraphics[width=0.21\textwidth]{1071_orig_512.jpg} \\
  \includegraphics[width=0.21\textwidth]{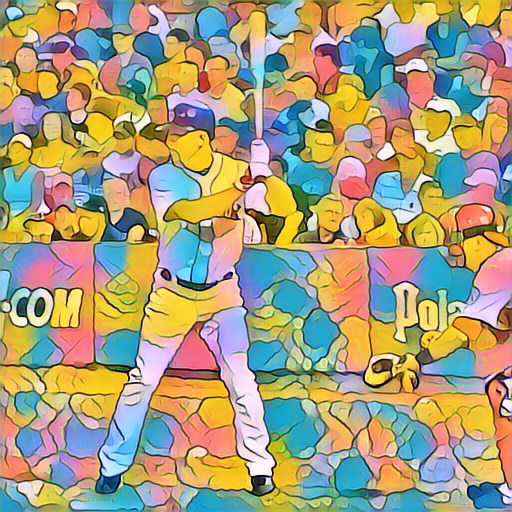}
  \includegraphics[width=0.21\textwidth]{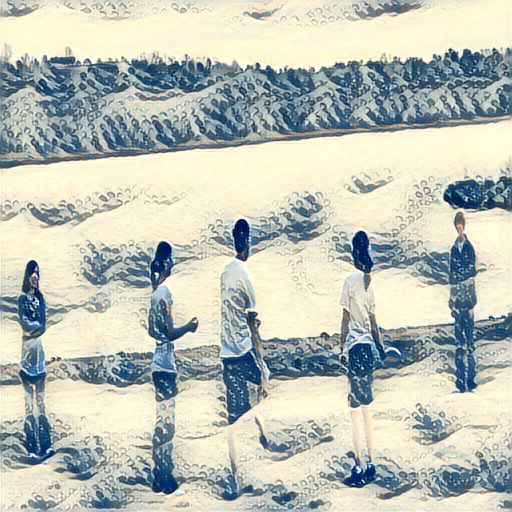}
  \includegraphics[width=0.21\textwidth]{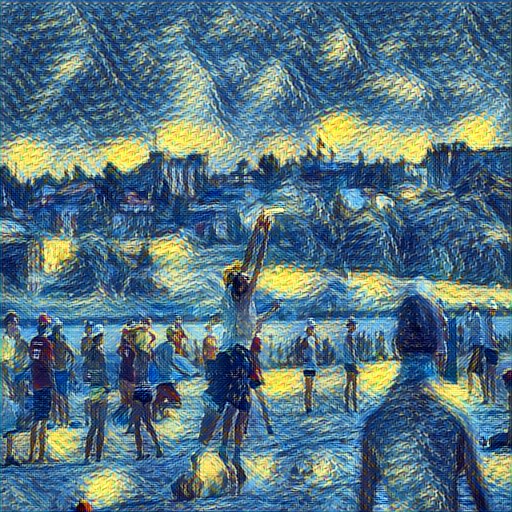}
  \includegraphics[width=0.21\textwidth]{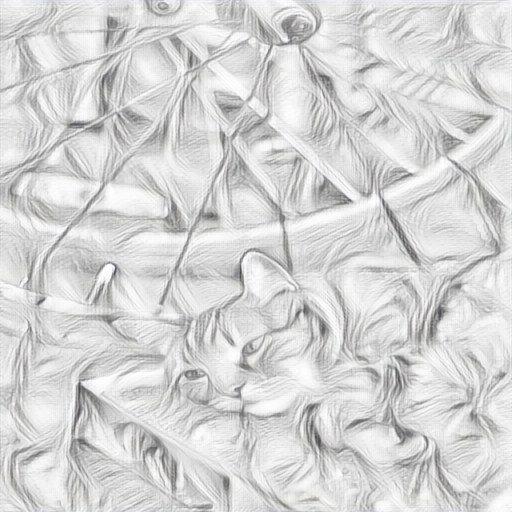}
  \vspace{-3mm}
  \caption{Example results for style transfer on $512\times512$ images. The model
    is applied in in a fully-convolutional manner to high-resolution images at test-time.
    The style images are the same as Figure~\ref{fig:style-results}.
  }
  \label{fig:style-results-512}
\end{figure}

Although our models are trained with $256\times256$ images, they can be applied in a fully-convolutional
manner to images of any size at test-time. In Figure~\ref{fig:style-results-512} we show examples
of style transfer using our models on $512\times512$ images.

In these results it is clear that the trained style transfer network is aware of the
\emph{semantic content} of images. For example in
the beach image in Figure~\ref{fig:style-results-512} the people are clearly recognizable in the
transformed image but the background is warped beyond recognition; similarly in the cat image,
the cat's face is clear in the transformed image, but its body is not. One
explanation is that the VGG-16 loss network has features which are selective for people and animals
since these objects are present in the classification dataset on which it was trained. Our style
transfer networks are trained to preserve VGG-16 features, and in doing so they learn to preserve
people and animals more than background objects.

\vspace{1mm}
\noindent \textbf{Quantitative Results.}
The baseline and our method both minimize Equation~\ref{eq:style}. The baseline performs explicit
optimization over the output image, while our method is trained to find a solution for any
content image $y_c$ in a single forward pass. We may therefore quantitatively compare the two
methods by measuring the degree to which they successfully minimize Equation~\ref{eq:style}.

We run our method and the baseline on 50 images from the MS-COCO validation set, using
\emph{The Muse} by Pablo Picasso as a style image. For the baseline we record the value of the
objective function at each iteration of optimization, and for our method we record the value
of Equation~\ref{eq:style} for each image; we also compute the value of Equation~\ref{eq:style}
when $y$ is equal to the content image $y_c$. Results are shown in Figure~\ref{fig:style-loss}.
We see that the content image $y_c$ achieves a very high loss, and that our method achieves a
loss comparable to between 50 and 100 iterations of explicit optimization.

Although our networks are trained to minimize Equation~\ref{eq:style} for $256\times256$ images,
they are also successful at minimizing the objective when applied to larger images. We repeat the
same quantitative evaluation for 50 images at $512\times512$ and $1024\times1024$; results are shown
in Figure~\ref{fig:style-loss}. We see that even at higher resolutions our model achieves a loss
comparable to 50 to 100 iterations of the baseline method.

\begin{table}[t]
  \centering
  \begin{tabular}{c||ccc|c||ccc}
    & \multicolumn{3}{c|}{Gatys \etal~\cite{gatys2015neural}} & 
    & \multicolumn{3}{c}{Speedup}\\
    Image Size & 100 & 300 & 500 \hspace{1mm} &
    \hspace{1mm} Ours \hspace{1mm} & \hspace{1mm} 100 & 300 & 500 \\
    \hline
    $256 \times 256$ & 3.17 & 9.52s & 15.86s &
    \textbf{0.015s} & 212x & 636x & \textbf{1060x} \\
    $512 \times 512$ & 10.97 & 32.91s & 54.85s &
    \textbf{0.05s} & 205x & 615x & \textbf{1026x} \\
    $1024 \times 1024$ & 42.89 & 128.66s & 214.44s &
    \textbf{0.21s} & 208x & 625x & \textbf{1042x}\\
  \end{tabular}
  \vspace{1mm}
  \caption{Speed (in seconds) for our style transfer network vs the optimization-based
    baseline for varying numbers of iterations and image resolutions.
    Our method gives similar qualitative results (see Figure~\ref{fig:style-results})
    but is faster than a single optimization step of the baseline method. Both methods
    are benchmarked on a GTX Titan X GPU.
  }
  \label{tab:style-speed}
  \vspace{-8mm}
\end{table}

\noindent \textbf{Speed.}
In Table~\ref{tab:style-speed} we compare the runtime of our method and the baseline
for several image sizes; for the baseline we report times for varying numbers of optimization
iterations. Across all image sizes, we see that the runtime of our method is approximately
twice the speed of a single iteration of the baseline method. Compared to 500 iterations of
the baseline method, our method is three orders of magnitude faster. Our method processes
images of size $512\times512$ at 20 FPS, making it feasible to run style transfer in
real-time or on video.

\def\mywidth{0.17\textwidth}
\begin{figure}[t]
  \centering
  \includegraphics[width=\mywidth]{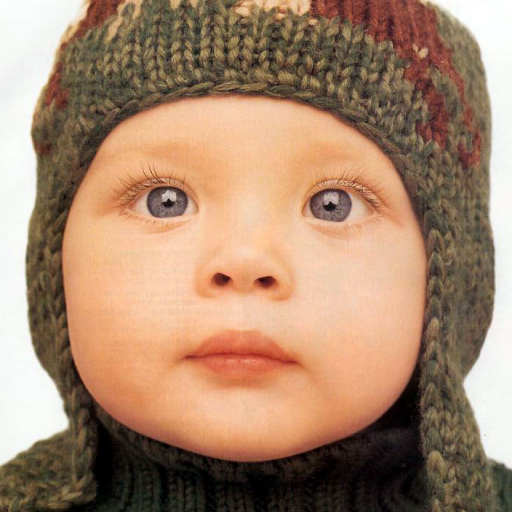}
  \includegraphics[width=\mywidth]{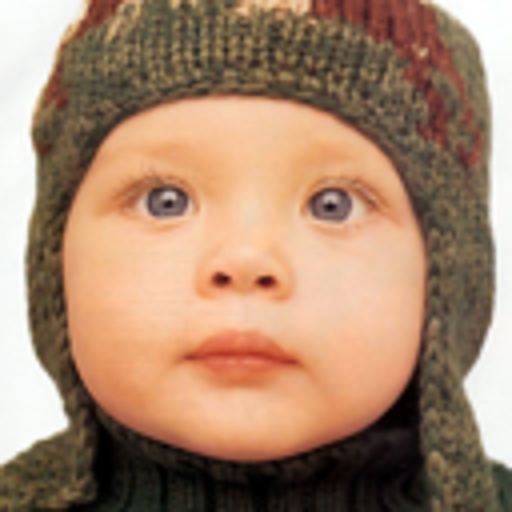}
  \includegraphics[width=\mywidth]{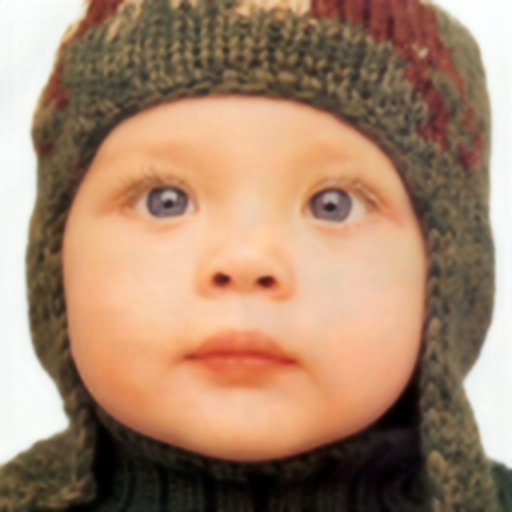}
  \includegraphics[width=\mywidth]{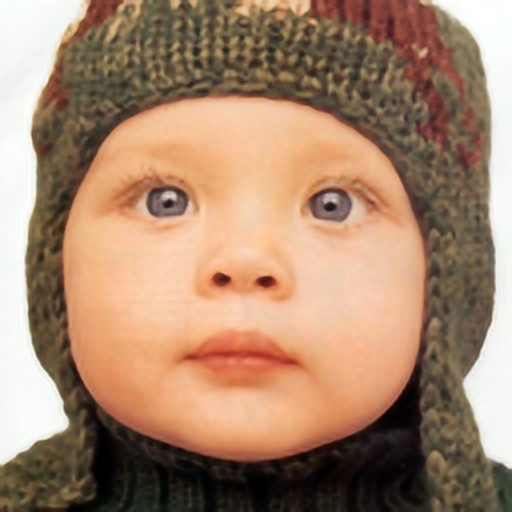}
  \includegraphics[width=\mywidth]{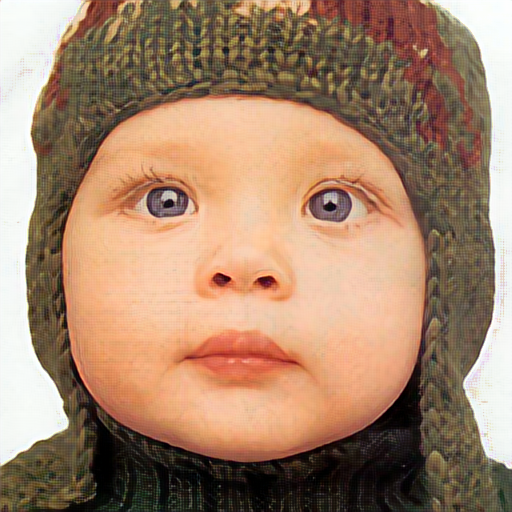} \\
  \includegraphics[trim={100px 280px 300px 160px},width=\mywidth,clip]{set5_1_orig.png}
  \includegraphics[trim={100px 280px 300px 160px},width=\mywidth,clip]{set5_1_bicubic.png}
  \includegraphics[trim={100px 280px 300px 160px},width=\mywidth,clip]{set5_1_pix.png}
  \includegraphics[trim={100px 280px 300px 160px},width=\mywidth,clip]{set5_1_srcnn.png}
  \includegraphics[trim={100px 280px 300px 160px},width=\mywidth,clip]{set5_1_feat.png} \\
  \vspace{1mm}
  \begin{minipage}{0.20\textwidth}
    \centering \textbf{Ground Truth} \\ This image \\ Set5 mean
  \end{minipage}
  \hspace{-0.02\textwidth}
  \begin{minipage}{\mywidth}
    \centering \textbf{Bicubic} \\ 31.78 / 0.8577 \\ 28.43 / 0.8114
  \end{minipage}
  \begin{minipage}{\mywidth}
    \centering \textbf{Ours} ($\ell_{pixel}$) \\ 31.47 / 0.8573 \\ 28.40 / 0.8205
  \end{minipage}
  \begin{minipage}{\mywidth}
    \centering \textbf{SRCNN}~\cite{dong2014learning} \\ 32.99 / 0.8784 \\ 30.48 / 0.8628
  \end{minipage}
  \begin{minipage}{\mywidth}
    \centering \textbf{Ours} ($\ell_{feat}$) \\ 29.24 / 0.7841 \\ 27.09 / 0.7680
  \end{minipage} \\
  \vspace{1mm}
  \includegraphics[width=\mywidth]{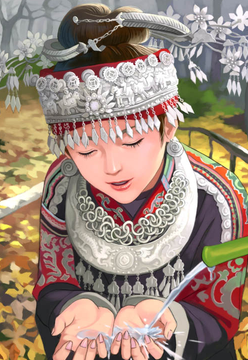}
  \includegraphics[width=\mywidth]{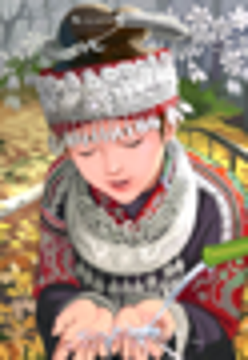}
  \includegraphics[width=\mywidth]{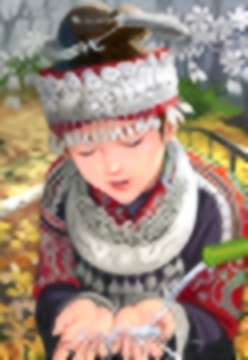}
  \includegraphics[width=\mywidth]{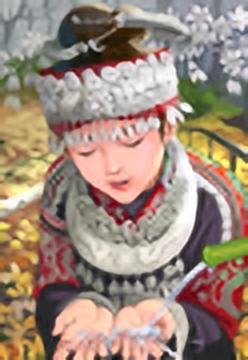}
  \includegraphics[width=\mywidth]{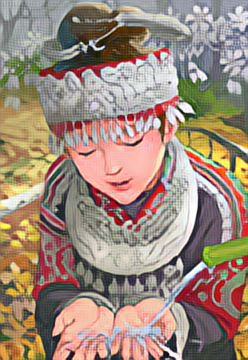} \\
  \includegraphics[trim={30px 200px 50px 100px},clip,width=\mywidth]{set14_5_orig.png}
  \includegraphics[trim={30px 200px 50px 100px},clip,width=\mywidth]{set14_5_bicubic.png}
  \includegraphics[trim={30px 200px 50px 100px},clip,width=\mywidth]{set14_5_pixel.png}
  \includegraphics[trim={30px 200px 50px 100px},clip,width=\mywidth]{set14_5_srcnn.png}
  \includegraphics[trim={30px 200px 50px 100px},clip,width=\mywidth]{set14_5_feat.png}
  \vspace{1mm}
  \begin{minipage}{0.20\textwidth}
    \centering \textbf{Ground Truth} \\ This Image \\ Set14 mean \\ BSD100 mean
  \end{minipage}
  \hspace{-0.02\textwidth}
  \begin{minipage}{\mywidth}
    \centering \textbf{Bicubic} \\ 21.69 / 0.5840 \\ 25.99 / 0.7301 \\ 25.96 / 0.682
  \end{minipage}
  \begin{minipage}{\mywidth}
    \centering \textbf{Ours} ($\ell_{pixel}$) \\ 21.66 / 0.5881 \\ 25.75 / 0.6994 \\ 25.91 / 0.6680
  \end{minipage}
  \begin{minipage}{\mywidth}
    \centering \textbf{SRCNN}~\cite{dong2014learning} \\ 22.53 / 0.6524 \\ 27.49 / 0.7503 \\ 
        26.90 / 0.7101
  \end{minipage}
  \begin{minipage}{\mywidth}
    \centering \textbf{Ours} ($\ell_{feat}$) \\  21.04 / 0.6116 \\ 24.99 / 0.6731 \\ 24.95 / 63.17
  \end{minipage}
  \vspace{-3mm}
  \caption{Results for $\times4$ super-resolution on images
    from Set5 (top) and Set14 (bottom). We report PSNR / SSIM for each example
    and the mean for each dataset. More results are shown in the supplementary material.
  }
  \label{fig:SR-4x-results}
  \vspace{-7mm}
\end{figure}

\vspace{-3mm}
\subsection{Single-Image Super-Resolution}
\label{sec:super-res}
In single-image super-resolution, the task is to generate a high-resolution output image from
a low-resolution input. This is an inherently ill-posed problem, since for each low-resolution
image there exist multiple high-resolution images that could have generated it. The ambiguity
becomes more extreme as the super-resolution factor grows; for large factors ($\times4$, $\times8$),
fine details of the high-resolution image may have little or no evidence in its low-resolution version.

%

To overcome this problem, we train super-resolution networks not with the per-pixel
loss typically used~\cite{dong2015image} but instead with a feature reconstruction
loss (see Section~\ref{sec:method}) to allow transfer of semantic knowledge from
the pretrained loss network to the super-resolution network. We focus on $\times4$ and $\times8$
super-resolution since larger factors require more semantic reasoning about the input.

The traditional metrics used to evaluate super-resolution are PSNR and SSIM~\cite{wang2004image},
both of which have been found to correlate poorly with human assessment of visual
quality~\cite{hanhart2013benchmarking,wang2009mean,huynh2008scope,sheikh2006statistical,kundu2015full}.
PSNR and SSIM rely only on low-level differences between pixels and operate under the
assumption of additive Gaussian noise, which may be invalid for super-resolution. In addition, PSNR
is equivalent to the per-pixel loss $\ell_{pixel}$, so as measured by PSNR a model trained to minimize
per-pixel loss should always outperform a model trained to minimize feature reconstruction loss.
We therefore emphasize that the goal of these experiments is not to achieve state-of-the-art PSNR or
SSIM results, but instead to showcase the qualitative difference between models trained with per-pixel
and feature reconstruction losses.

\vspace{1mm}
\noindent \textbf{Model Details.}
We train models to perform $\times4$ and $\times8$ super-resolution by minimizing feature
reconstruction loss at layer \verb.relu2_2. from the VGG-16 loss network $\phi$. We train with
$288\times288$ patches from 10k images from the MS-COCO training set, and prepare low-resolution
inputs by blurring with a Gaussian kernel of width $\sigma=1.0$ and downsampling with bicubic
interpolation. We train with a batch size of 4 for 200k iterations using Adam~\cite{kingma2014adam}
with a learning rate of $1\times10^{-3}$ without weight decay or dropout. As a post-processing step,
we perform histogram matching between our network output and the low-resolution input.

\vspace{1mm}
\noindent \textbf{Baselines.}
As a baseline model we use SRCNN~\cite{dong2015image} for its state-of-the-art performance. SRCNN
is a three-layer convolutional network trained to minimize per-pixel loss on $33\times33$
patches from the ILSVRC 2013 detection dataset. SRCNN is not trained for $\times8$ super-resolution,
so we can only evaluate it on $\times4$.

SRCNN is trained for more than $10^9$ iterations, which is not computationally feasible for our
models. To account for differences between SRCNN and our model in data, training, and architecture,
we train image transformation networks for $\times4$ and $\times8$ super-resolution using
$\ell_{pixel}$; these networks use identical data, architecture, and training as the networks trained
to minimize $\ell_{feat}$.


\vspace{1mm}
\noindent \textbf{Evaluation.}
We evaluate all models on the standard Set5~\cite{bevilacqua2012low},
Set14~\cite{zeyde2010single}, and BSD100~\cite{huang2015single} datasets. We report PSNR and
SSIM~\cite{wang2004image}, computing both only on the Y channel after converting to the YCbCr
colorspace, following~\cite{dong2015image,timofte2014adjusted}.


\vspace{1mm}
\noindent \textbf{Results.}
We show results for $\times4$ super-resolution in Figure~\ref{fig:SR-4x-results}.
Compared to the other methods, our model trained for feature reconstruction does a very good job
at reconstructing sharp edges and fine details, such as the eyelashes in the first image and the
individual elements of the hat in the second image. The feature reconstruction loss gives rise to
a slight cross-hatch pattern visible under magnification, which harms its PSNR and SSIM compared
to baseline methods.

\begin{figure}[t]
   \centering
  \includegraphics[width=0.22\textwidth]{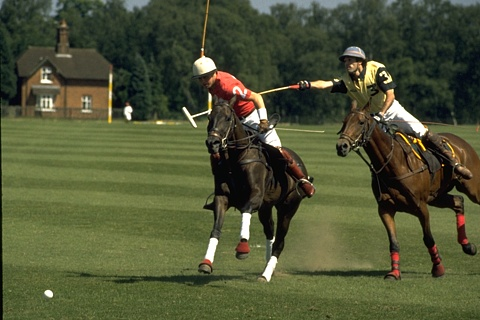}
  \includegraphics[width=0.22\textwidth]{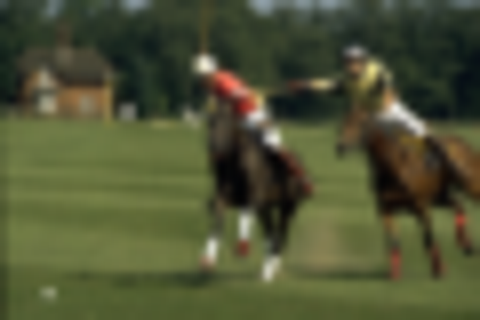}
  \includegraphics[width=0.22\textwidth]{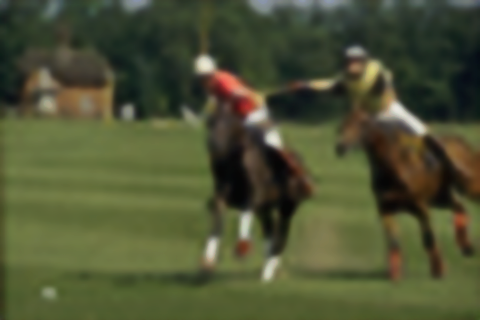}
  \includegraphics[width=0.22\textwidth]{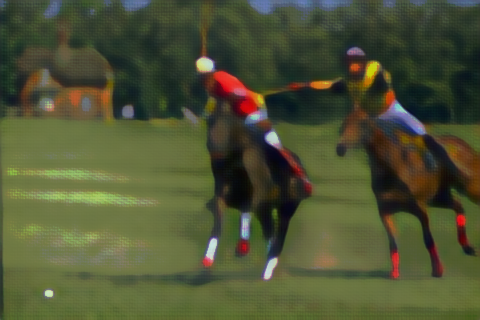} \\
  \includegraphics[trim={160px 30px 150px 200px},width=0.22\textwidth,clip]{bsd070_orig.png}
  \includegraphics[trim={160px 30px 150px 200px},width=0.22\textwidth,clip]{bsd070_bicubic.png}
  \includegraphics[trim={160px 30px 150px 200px},width=0.22\textwidth,clip]{bsd070_pix.png}
  \includegraphics[trim={160px 30px 150px 200px},width=0.22\textwidth,clip]{bsd070_feat.png} \\
  \vspace{1mm}
  \begin{minipage}{0.22\textwidth}
    \centering \textbf{Ground Truth} \\ This image \\ Set5 mean \\ Set14 mean \\ BSD100 mean
  \end{minipage}
  \begin{minipage}{0.22\textwidth}
    \centering \textbf{Bicubic} \\ 22.75 / 0.5946 \\ 23.80 / 0.6455 \\ 22.37 / 0.5518 \\ 22.11 / 0.5322
  \end{minipage}
  \begin{minipage}{0.22\textwidth}
    \centering \textbf{Ours} ($\ell_{pixel}$) \\ 23.42 / 0.6168 \\ 24.77 / 0.6864 \\ 23.02 / 0.5787
      \\ 22.54 / 0.5526 
  \end{minipage}
  \begin{minipage}{0.22\textwidth}
    \centering \textbf{Ours} ($\ell_{feat}$) \\  21.90 / 0.6083 \\ 23.26 / 0.7058 \\ 21.64 / 0.5837 
      \\ 21.35 / 0.5474
  \end{minipage} \\
  \vspace{-2mm}
  \caption{Super-resolution results with scale factor $\times8$ on an image from the
    BSD100 dataset. We report PSNR / SSIM for the example image and the mean
    for each dataset. More results are shown in the supplementary material.
  }
  \vspace{-4mm}
  \label{fig:SR-8x-results}
\end{figure}

Results for $\times8$ super-resolution are shown in Figure~\ref{fig:SR-8x-results}. Again we see that
our $\ell_{feat}$ model does a good job at edges and fine details compared to other models,
such as the horse's legs and hooves. The $\ell_{feat}$ model does not sharpen edges
indiscriminately; compared to the $\ell_{pixel}$ model, the $\ell_{feat}$ model sharpens the boundary
edges of the horse and rider but the background trees remain diffuse, suggesting that the
$\ell_{feat}$ model may be more aware of image semantics.

Since our $\ell_{pixel}$ and our $\ell_{feat}$ models share the same architecture,
data, and training procedure, all differences between them are due to the difference between the
$\ell_{pixel}$ and $\ell_{feat}$ losses. The $\ell_{pixel}$ loss gives fewer visual artifacts and
higher PSNR values but the $\ell_{feat}$ loss does a better job at reconstructing fine details, leading
to pleasing visual results.


\vspace{-2mm}
\section{Conclusion}
\vspace{-2mm}
In this paper we have combined the benefits of feed-forward image transformation tasks and
optimization-based methods for image generation by training feed-forward transformation networks
with perceptual loss functions. We have applied this method to style transfer where we achieve
comparable performance and drastically improved speed compared to existing methods, and to
single-image super-resolution where we show that training with a perceptual loss allows the model
to better reconstruct fine details and edges.

In future work we hope to explore the use of perceptual loss functions for other image transformation
tasks, such as colorization and semantic segmentation. We also plan to investigate the use of different
loss networks to see whether for example loss networks trained on different tasks or datasets can
impart image transformation networks with different types of semantic knowledge.

\clearpage
\bibliographystyle{splncs}
\bibliography{eccv16} 

\begin{thebibliography}{10}

\bibitem{dong2015image}
Dong, C., Loy, C.C., He, K., Tang, X.:
\newblock Image super-resolution using deep convolutional networks.
\newblock (2015)

\bibitem{cheng2015deep}
Cheng, Z., Yang, Q., Sheng, B.:
\newblock Deep colorization.
\newblock In: Proceedings of the IEEE International Conference on Computer
  Vision. (2015)  415--423

\bibitem{long_shelhamer_fcn}
Long, J., Shelhamer, E., Darrell, T.:
\newblock Fully convolutional networks for semantic segmentation.
\newblock CVPR (2015)

\bibitem{eigen2014depth}
Eigen, D., Puhrsch, C., Fergus, R.:
\newblock Depth map prediction from a single image using a multi-scale deep
  network.
\newblock In: Advances in Neural Information Processing Systems. (2014)
  2366--2374

\bibitem{eigen2015predicting}
Eigen, D., Fergus, R.:
\newblock Predicting depth, surface normals and semantic labels with a common
  multi-scale convolutional architecture.
\newblock In: Proceedings of the IEEE International Conference on Computer
  Vision. (2015)  2650--2658

\bibitem{mahendran15understanding}
Mahendran, A., Vedaldi, A.:
\newblock Understanding deep image representations by inverting them.
\newblock In: Proceedings of the {IEEE} Conf. on Computer Vision and Pattern
  Recognition ({CVPR}). (2015)

\bibitem{simonyan2013deep}
Simonyan, K., Vedaldi, A., Zisserman, A.:
\newblock Deep inside convolutional networks: Visualising image classification
  models and saliency maps.
\newblock arXiv preprint arXiv:1312.6034 (2013)

\bibitem{yosinski2015understanding}
Yosinski, J., Clune, J., Nguyen, A., Fuchs, T., Lipson, H.:
\newblock Understanding neural networks through deep visualization.
\newblock arXiv preprint arXiv:1506.06579 (2015)

\bibitem{Gatys2015b}
Gatys, L.A., Ecker, A.S., Bethge, M.:
\newblock Texture synthesis using convolutional neural networks.
\newblock In: Advances in Neural Information Processing Systems 28. (May 2015)

\bibitem{gatys2015neural}
Gatys, L.A., Ecker, A.S., Bethge, M.:
\newblock A neural algorithm of artistic style.
\newblock arXiv preprint arXiv:1508.06576 (2015)

\bibitem{dong2014learning}
Dong, C., Loy, C.C., He, K., Tang, X.:
\newblock Learning a deep convolutional network for image super-resolution.
\newblock In: Computer Vision--ECCV 2014.
\newblock Springer (2014)  184--199

\bibitem{farabet2013learning}
Farabet, C., Couprie, C., Najman, L., LeCun, Y.:
\newblock Learning hierarchical features for scene labeling.
\newblock Pattern Analysis and Machine Intelligence, IEEE Transactions on
  \textbf{35}(8) (2013)  1915--1929

\bibitem{pinheiro2013recurrent}
Pinheiro, P.H., Collobert, R.:
\newblock Recurrent convolutional neural networks for scene parsing.
\newblock arXiv preprint arXiv:1306.2795 (2013)

\bibitem{noh2015learning}
Noh, H., Hong, S., Han, B.:
\newblock Learning deconvolution network for semantic segmentation.
\newblock arXiv preprint arXiv:1505.04366 (2015)

\bibitem{zheng2015conditional}
Zheng, S., Jayasumana, S., Romera-Paredes, B., Vineet, V., Su, Z., Du, D.,
  Huang, C., Torr, P.H.:
\newblock Conditional random fields as recurrent neural networks.
\newblock In: Proceedings of the IEEE International Conference on Computer
  Vision. (2015)  1529--1537

\bibitem{liu2015deep}
Liu, F., Shen, C., Lin, G.:
\newblock Deep convolutional neural fields for depth estimation from a single
  image.
\newblock In: Proceedings of the IEEE Conference on Computer Vision and Pattern
  Recognition. (2015)  5162--5170

\bibitem{wang2015designing}
Wang, X., Fouhey, D., Gupta, A.:
\newblock Designing deep networks for surface normal estimation.
\newblock In: Proceedings of the IEEE Conference on Computer Vision and Pattern
  Recognition. (2015)  539--547

\bibitem{szegedy2013intriguing}
Szegedy, C., Zaremba, W., Sutskever, I., Bruna, J., Erhan, D., Goodfellow, I.,
  Fergus, R.:
\newblock Intriguing properties of neural networks.
\newblock arXiv preprint arXiv:1312.6199 (2013)

\bibitem{nguyen2015deep}
Nguyen, A., Yosinski, J., Clune, J.:
\newblock Deep neural networks are easily fooled: High confidence predictions
  for unrecognizable images.
\newblock In: Computer Vision and Pattern Recognition (CVPR), 2015 IEEE
  Conference on, IEEE (2015)  427--436

\bibitem{d2012beyond}
d'Angelo, E., Alahi, A., Vandergheynst, P.:
\newblock Beyond bits: Reconstructing images from local binary descriptors.
\newblock In: Pattern Recognition (ICPR), 2012 21st International Conference
  on, IEEE (2012)  935--938

\bibitem{vondrick2013hoggles}
Vondrick, C., Khosla, A., Malisiewicz, T., Torralba, A.:
\newblock Hoggles: Visualizing object detection features.
\newblock In: Proceedings of the IEEE International Conference on Computer
  Vision. (2013)  1--8

\bibitem{dosovitskiy2015inverting}
Dosovitskiy, A., Brox, T.:
\newblock Inverting visual representations with convolutional networks.
\newblock arXiv preprint arXiv:1506.02753 (2015)

\bibitem{yang2014single}
Yang, C.Y., Ma, C., Yang, M.H.:
\newblock Single-image super-resolution: a benchmark.
\newblock In: Computer Vision--ECCV 2014.
\newblock Springer (2014)  372--386

\bibitem{irani1991improving}
Irani, M., Peleg, S.:
\newblock Improving resolution by image registration.
\newblock CVGIP: Graphical models and image processing \textbf{53}(3) (1991)
  231--239

\bibitem{freedman2011image}
Freedman, G., Fattal, R.:
\newblock Image and video upscaling from local self-examples.
\newblock ACM Transactions on Graphics (TOG) \textbf{30}(2) (2011) ~12

\bibitem{sun2008image}
Sun, J., Sun, J., Xu, Z., Shum, H.Y.:
\newblock Image super-resolution using gradient profile prior.
\newblock In: Computer Vision and Pattern Recognition, 2008. CVPR 2008. IEEE
  Conference on, IEEE (2008)  1--8

\bibitem{shan2008fast}
Shan, Q., Li, Z., Jia, J., Tang, C.K.:
\newblock Fast image/video upsampling.
\newblock In: ACM Transactions on Graphics (TOG). Volume~27., ACM (2008)  153

\bibitem{kim2010single}
Kim, K.I., Kwon, Y.:
\newblock Single-image super-resolution using sparse regression and natural
  image prior.
\newblock Pattern Analysis and Machine Intelligence, IEEE Transactions on
  \textbf{32}(6) (2010)  1127--1133

\bibitem{xiong2010robust}
Xiong, Z., Sun, X., Wu, F.:
\newblock Robust web image/video super-resolution.
\newblock Image Processing, IEEE Transactions on \textbf{19}(8) (2010)
  2017--2028

\bibitem{freeman2002example}
Freeman, W.T., Jones, T.R., Pasztor, E.C.:
\newblock Example-based super-resolution.
\newblock Computer Graphics and Applications, IEEE \textbf{22}(2) (2002)
  56--65

\bibitem{chang2004super}
Chang, H., Yeung, D.Y., Xiong, Y.:
\newblock Super-resolution through neighbor embedding.
\newblock In: Computer Vision and Pattern Recognition, 2004. CVPR 2004.
  Proceedings of the 2004 IEEE Computer Society Conference on. Volume~1., IEEE
  (2004)  I--I

\bibitem{glasner2009super}
Glasner, D., Bagon, S., Irani, M.:
\newblock Super-resolution from a single image.
\newblock In: Computer Vision, 2009 IEEE 12th International Conference on, IEEE
  (2009)  349--356

\bibitem{yang2013fast}
Yang, J., Lin, Z., Cohen, S.:
\newblock Fast image super-resolution based on in-place example regression.
\newblock In: Proceedings of the IEEE Conference on Computer Vision and Pattern
  Recognition. (2013)  1059--1066

\bibitem{sun2003image}
Sun, J., Zheng, N.N., Tao, H., Shum, H.Y.:
\newblock Image hallucination with primal sketch priors.
\newblock In: Computer Vision and Pattern Recognition, 2003. Proceedings. 2003
  IEEE Computer Society Conference on. Volume~2., IEEE (2003)  II--729

\bibitem{ni2007image}
Ni, K.S., Nguyen, T.Q.:
\newblock Image superresolution using support vector regression.
\newblock Image Processing, IEEE Transactions on \textbf{16}(6) (2007)
  1596--1610

\bibitem{he2013beta}
He, L., Qi, H., Zaretzki, R.:
\newblock Beta process joint dictionary learning for coupled feature spaces
  with application to single image super-resolution.
\newblock In: Proceedings of the IEEE Conference on Computer Vision and Pattern
  Recognition. (2013)  345--352

\bibitem{yang2008image}
Yang, J., Wright, J., Huang, T., Ma, Y.:
\newblock Image super-resolution as sparse representation of raw image patches.
\newblock In: Computer Vision and Pattern Recognition, 2008. CVPR 2008. IEEE
  Conference on, IEEE (2008)  1--8

\bibitem{yang2010image}
Yang, J., Wright, J., Huang, T.S., Ma, Y.:
\newblock Image super-resolution via sparse representation.
\newblock Image Processing, IEEE Transactions on \textbf{19}(11) (2010)
  2861--2873

\bibitem{timofte2014adjusted}
Timofte, R., De~Smet, V., Van~Gool, L.:
\newblock A+: Adjusted anchored neighborhood regression for fast
  super-resolution.
\newblock In: Computer Vision--ACCV 2014.
\newblock Springer (2014)  111--126

\bibitem{schulter2015fast}
Schulter, S., Leistner, C., Bischof, H.:
\newblock Fast and accurate image upscaling with super-resolution forests.
\newblock In: Proceedings of the IEEE Conference on Computer Vision and Pattern
  Recognition. (2015)  3791--3799

\bibitem{huang2015single}
Huang, J.B., Singh, A., Ahuja, N.:
\newblock Single image super-resolution from transformed self-exemplars.
\newblock In: Computer Vision and Pattern Recognition (CVPR), 2015 IEEE
  Conference on, IEEE (2015)  5197--5206

\bibitem{radford2015unsupervised}
Radford, A., Metz, L., Chintala, S.:
\newblock Unsupervised representation learning with deep convolutional
  generative adversarial networks.
\newblock arXiv preprint arXiv:1511.06434 (2015)

\bibitem{he2015deep}
He, K., Zhang, X., Ren, S., Sun, J.:
\newblock Deep residual learning for image recognition.
\newblock arXiv preprint arXiv:1512.03385 (2015)

\bibitem{gross2016training}
Gross, S., Wilber, M.:
\newblock Training and investigating residual nets.
\newblock \url{http://torch.ch/blog/2016/02/04/resnets.html} (2016)

\bibitem{ioffe2015batch}
Ioffe, S., Szegedy, C.:
\newblock Batch normalization: Accelerating deep network training by reducing
  internal covariate shift.
\newblock In: Proceedings of The 32nd International Conference on Machine
  Learning. (2015)  448--456

\bibitem{simonyan2014very}
Simonyan, K., Zisserman, A.:
\newblock Very deep convolutional networks for large-scale image recognition.
\newblock arXiv preprint arXiv:1409.1556 (2014)

\bibitem{ILSVRC15}
Russakovsky, O., Deng, J., Su, H., Krause, J., Satheesh, S., Ma, S., Huang, Z.,
  Karpathy, A., Khosla, A., Bernstein, M., Berg, A.C., Fei-Fei, L.:
\newblock {ImageNet Large Scale Visual Recognition Challenge}.
\newblock International Journal of Computer Vision (IJCV) \textbf{115}(3)
  (2015)  211--252

\bibitem{aly2005image}
Aly, H.A., Dubois, E.:
\newblock Image up-sampling using total-variation regularization with a new
  observation model.
\newblock Image Processing, IEEE Transactions on \textbf{14}(10) (2005)
  1647--1659

\bibitem{zhang2010non}
Zhang, H., Yang, J., Zhang, Y., Huang, T.S.:
\newblock Non-local kernel regression for image and video restoration.
\newblock In: Computer Vision--ECCV 2010.
\newblock Springer (2010)  566--579

\bibitem{lin2014microsoft}
Lin, T.Y., Maire, M., Belongie, S., Hays, J., Perona, P., Ramanan, D.,
  Doll{\'a}r, P., Zitnick, C.L.:
\newblock Microsoft coco: Common objects in context.
\newblock In: Computer Vision--ECCV 2014.
\newblock Springer (2014)  740--755

\bibitem{kingma2014adam}
Kingma, D., Ba, J.:
\newblock Adam: A method for stochastic optimization.
\newblock arXiv preprint arXiv:1412.6980 (2014)

\bibitem{collobert2011torch7}
Collobert, R., Kavukcuoglu, K., Farabet, C.:
\newblock Torch7: A matlab-like environment for machine learning.
\newblock In: BigLearn, NIPS Workshop. Number EPFL-CONF-192376 (2011)

\bibitem{chetlur2014cudnn}
Chetlur, S., Woolley, C., Vandermersch, P., Cohen, J., Tran, J., Catanzaro, B.,
  Shelhamer, E.:
\newblock cudnn: Efficient primitives for deep learning.
\newblock arXiv preprint arXiv:1410.0759 (2014)

\bibitem{wang2004image}
Wang, Z., Bovik, A.C., Sheikh, H.R., Simoncelli, E.P.:
\newblock Image quality assessment: from error visibility to structural
  similarity.
\newblock Image Processing, IEEE Transactions on \textbf{13}(4) (2004)
  600--612

\bibitem{hanhart2013benchmarking}
Hanhart, P., Korshunov, P., Ebrahimi, T.:
\newblock Benchmarking of quality metrics on ultra-high definition video
  sequences.
\newblock In: Digital Signal Processing (DSP), 2013 18th International
  Conference on, IEEE (2013)  1--8

\bibitem{wang2009mean}
Wang, Z., Bovik, A.C.:
\newblock Mean squared error: love it or leave it? a new look at signal
  fidelity measures.
\newblock Signal Processing Magazine, IEEE \textbf{26}(1) (2009)  98--117

\bibitem{huynh2008scope}
Huynh-Thu, Q., Ghanbari, M.:
\newblock Scope of validity of psnr in image/video quality assessment.
\newblock Electronics letters \textbf{44}(13) (2008)  800--801

\bibitem{sheikh2006statistical}
Sheikh, H.R., Sabir, M.F., Bovik, A.C.:
\newblock A statistical evaluation of recent full reference image quality
  assessment algorithms.
\newblock Image Processing, IEEE Transactions on \textbf{15}(11) (2006)
  3440--3451

\bibitem{kundu2015full}
Kundu, D., Evans, B.L.:
\newblock Full-reference visual quality assessment for synthetic images: A
  subjective study.
\newblock Proc. IEEE Int. Conf. on Image Processing (2015)

\bibitem{bevilacqua2012low}
Bevilacqua, M., Roumy, A., Guillemot, C., Alberi-Morel, M.L.:
\newblock Low-complexity single-image super-resolution based on nonnegative
  neighbor embedding.
\newblock (2012)

\bibitem{zeyde2010single}
Zeyde, R., Elad, M., Protter, M.:
\newblock On single image scale-up using sparse-representations.
\newblock In: Curves and Surfaces.
\newblock Springer (2010)  711--730

\end{thebibliography}
\end{document}